\documentclass{article}
\usepackage{arxiv}

\usepackage[utf8]{inputenc} 
\usepackage[T1]{fontenc}    
\usepackage{hyperref}       
\usepackage{url}            
\usepackage{booktabs}       
\usepackage{amsfonts}       
\usepackage{nicefrac}       
\usepackage{microtype}      
\usepackage{lipsum}

\usepackage{graphicx}
\usepackage{amssymb}
\usepackage{amsthm}
\usepackage{amsmath}

\usepackage[numbers,sort&compress ]{natbib}
\usepackage{float}
\usepackage[table]{xcolor}
\usepackage{xcolor}
\usepackage{pdflscape}
\usepackage{multirow}
\usepackage{subcaption}
\usepackage{array}
\newcolumntype{C}[1]{>{\centering\arraybackslash}p{#1}}

\title{PLANesT-3D: A new annotated dataset for segmentation of 3D plant point clouds}

\author{
  Kerem Mertoğlu \\
  Faculty of Agriculture\\
  Usak University\\
  Usak, Turkey \\
   \And
  Yusuf Şalk, Server Karahan Sarıkaya \\
  Department of Electrical-Electronics Engineering\\
  Eskisehir Osmangazi University\\
  Eskisehir, Turkey \\
  \And
  Kaya Turgut \\
  SADC Department\\
  Huawei Türkiye R\&D Center\\
  Istanbul, Turkey \\
  \texttt{kaya.turgut1@huawei.com}
   \And
  Yasemin Evrenesoğlu \\
  Faculty of Agriculture\\
  Eskisehir Osmangazi University\\
  Eskisehir, Turkey \\
  \And
  Hakan Çevikalp \\
  Department of Electrical-Electronics Engineering\\
  Eskisehir Osmangazi University\\
  Eskisehir, Turkey \\
  \And
  Ömer Nezih Gerek \\
  Department of Electrical-Electronics Engineering\\
  Eskisehir Osmangazi University\\
  Eskisehir, Turkey \\
  \And   
  Helin Dutağacı \\
  Department of Electrical-Electronics Engineering\\
  Eskisehir Osmangazi University\\
  Eskisehir, Turkey \\
  \texttt{hdutagaci@ogu.edu.tr} 
  \And
  David Rousseau \\
  Laboratoire Angevin de Recherche en Ingénierie des Systèmes (LARIS)\\
  Université d'Angers\\
  Angers, France \\ }

\begin{document}
\maketitle

\begin{abstract}
Creation of new annotated public datasets is crucial in helping advances in 3D computer vision and machine learning meet their full potential for automatic interpretation of 3D plant models.  Despite the proliferation of deep neural network architectures for segmentation and phenotyping of 3D plant models in the last decade, the amount of data, and diversity in terms of species and data acquisition modalities are far from sufficient for evaluation of such tools for their generalization ability. To contribute to closing this gap, we introduce PLANesT-3D; a new annotated dataset of 3D color point clouds of plants. PLANesT-3D is composed of 34 point cloud models representing 34 real plants from three different plant species: \textit{Capsicum annuum}, \textit{Rosa kordana}, and \textit{Ribes rubrum}. Both semantic labels in terms of "leaf" and "stem", and organ instance labels were manually annotated for the full point clouds. PLANesT-3D introduces diversity to existing datasets by adding point clouds of two new species and providing 3D data acquired with the low-cost SfM/MVS technique as opposed to laser scanning or expensive setups. Point clouds reconstructed with SfM/MVS modality exhibit challenges such as missing data, variable density, and illumination variations. As an additional contribution, SP-LSCnet, a novel semantic segmentation method that is a combination of unsupervised superpoint extraction and a 3D point-based deep learning approach is introduced and evaluated on the new dataset. The advantages of SP-LSCnet over other deep learning methods are its modular structure and increased interpretability. Two existing deep neural network architectures, PointNet++ and RoseSegNet, were also tested on the point clouds of PLANesT-3D for semantic segmentation. 
\end{abstract}

\keywords{3D point cloud \and 3D plant modeling \and 3D plant dataset \and 3D semantic segmentation}

\section{Introduction}
\label{section:intro}
The cultivation of horticultural crops is one of the most labor intensive economic activities and is central to global food security
\citep{Riisgaard2011, van2016horticultural}. The accuracy and efficiency of horticultural practices, such as pruning, weed control, disease management, and harvesting, directly influence the yield and quality of crops \citep{tyagi2017pre}. The increasing global demand for high-quality crops, and hence the demand for intensive labor, motivates the development of automation technologies in horticulture \citep{Bac2014, bogue2016robots}. Automated horticultural practices are expected to adapt to plant traits, which are diverse across plant species \citep{ZHANG2023, vanHenten2022}. Automation of trait measurements is also key to high throughput plant phenotyping, a fundamental process in plant research, breeding, and crop management \citep{atefi2021,Pieruschka2019,James2022}. Modeling of traits within a population and capturing the structure of individual plants or plant communities for particular operations have become major applications of computer vision \citep{Minervini2015,KOLHAR2023,Coleman2023, Zieschank2023community}.

3D computer vision is particularly relevant in many automated horticultural operations. Robotic systems designed to perform pruning, weeding, and harvesting require correct identification of plant parts and accurate estimation of the location, size, shape, and orientation of the structures in 3D space \citep{kootstra2021,RONG2022,ZAHID2021,LELOUEDEC2021}. Advances in 3D sensing technologies and machine learning fuel the development of algorithms for automatic segmentation of plants into their structural parts, as well as organ-level trait estimation \citep{gibbs2016approaches,paulus2019measuring}.

Despite the considerable progress in 3D geometry modeling and analysis of plants in the last decades \citep{Okura2022, harandi2023make}, creation of benchmark datasets and provision of publicly available annotated data for training machine learning techniques, which are fundamental components for advancement of relevant research, are still lagging behind compared to other domains \citep{BARTH2018,Chaudhury2020}. Annotated public datasets of plant models are of particular importance both for comparison of competing algorithms and also for providing training data to machine learning processes \citep{Coleman2023}.

\begin{table}[!ht]
  \centering
  \caption{Public datasets containing full 3D models of plants in comparison with the PLANesT-3D dataset introduced in this work.}
  \label{tab:datasets}
  
  \begin{tabular}{cccccc}
    \multirow{2}{*}{\textbf{dataset}} & \multicolumn{5}{c}{\textbf{Properties}} \\
    \cline{2-6}
    & \textbf{\# models}  & \textbf{Species} & \textbf{Modality} & \textbf{Color} & \textbf{Labels}\\
   \hline
   \textbf{Pheno4D}   & \multirow{2}{*}{126}  & Tomato & \multirow{2}{*}{Laser} & \multirow{2}{*}{No} & Semantic \& \\
	\cite{Schunck2021pheno4d} &  & Maize &  &  & Instance\\
    \hline
    \textbf{ROSE-X}  & \multirow{2}{*}{11}  & \multirow{2}{*}{Rose} & \multirow{2}{*}{X-rays}  & \multirow{2}{*}{No} & \multirow{2}{*}{Semantic} \\
    \cite{Dutagaci2020rose} & & & & \\
    \hline
     \textbf{Plant3D} & \multirow{4}{*}{714}  & Tomato & \multirow{4}{*}{Laser}  & \multirow{4}{*}{No} & \\\
 \cite{Conn2017high} & & Tobacco  & &  & Growth\\
 \cite{Conn2017statistical} & & Sorghum & &  & conditions\\
 \cite{Conn2019} & & Arabidopsis  & &  & \\
 \hline
 \textbf{Soybean-MVS}   & \multirow{2}{*}{102}  & \multirow{2}{*}{Soybean} & \multirow{2}{*}{MVS} & \multirow{2}{*}{Yes} & Semantic \&  \\
	\cite{Sun2023-Soybean} &  &  &  &  & Instance \\

       \hline
  \multirow{3}{*}{\textbf{PLANesT-3D}}  & \multirow{3}{*}{34}  & Pepper &   & \multirow{3}{*}{Yes} &  Semantic \\
  & & Rose  &  SfM-MVS  &  &  \& \\
  & & Ribes  &  &  &  Instance \\
    \hline
  \end{tabular}
\end{table}
In the last decade, open datasets of annotated 2D plant images have been proliferated to assist evaluation of performance and development of machine learning models for plant phenotyping \citep{LU2020dataSurvey}. However, open datasets containing annotated and complete 3D plant models are rare. Table \ref{tab:datasets} gives a list of such datasets. Plant3D \citep{Conn2017high,Conn2017statistical,Conn2019} contains a total of 714 3D laser scans of Tomato, Tobacco, Sorghum, and Arabidopsis plants obtained within 20-30 days of development. Information on the growth conditions of the plants is also provided. The 3D models do not contain color information. \cite{Ziamtsov2019} manually labeled 54 models of the Plant3D data into stem and lamina points, and evaluated a range of machine learning approaches on this subset. \cite{Li2022PSegNet} annotated 546 of the plants of the Plant3D dataset with semantic and instance labels. The ROSE-X data set \citep{Dutagaci2020rose}
contains 3D models of 11 rosebush plants acquired by X-ray computer tomography. The voxels in the volumetric models were labeled into three semantic categories: "Leaf", "Stem", and "Flower". The point cloud versions of the volumetric models are also available. Pheno4D \citep{Schunck2021pheno4d} is a dataset of 3D point clouds of 7 maize plants and 7 tomato plants. The plants were scanned with a laser scanner at different stages of growth, resulting in 244 point clouds. 126 of them were manually annotated with semantic and instance labels. The Soybean-MVS dataset \citep{Sun2023-Soybean} is fundamentally different from the rest in terms of data acquisition modality. The plants were captured with an RGB camera in a controlled setup, and their corresponding point clouds were created through multi-view stereo. A total of 102 point cloud models of five different soybean varieties were reconstructed at 13 stages of the whole growth period.

These 3D plant datasets have been often used for assessing off-the-shelf machine learning tools for semantic and instance segmentation \citep{Ziamtsov2019, Schunck2021pheno4d, Turgut2022segmentation,Sun2023-Soybean}. The datasets also enabled training and evaluation of new approaches specifically developed for plant data \citep{Ziamtsov2020,mirande2022,Xiang2022research, dutagaci2023using}, including deep learning architectures such as PSegNet \citep{Li2022PSegNet}, PlantNet \citep{Li2022plantnet}, RoseSegNet \citep{Turgut2022rosesegnet}, and FF-Net \citep{Guo2023FFnet}.

Although smaller in terms of the number of models as compared to Pheno4D, Plant3D, and Soybean-MVS (Table \ref{tab:datasets}), the new dataset (PLANesT-3D \footnote{The PLANesT-3D dataset is publicly available at \url{https://aperta.ulakbim.gov.tr/record/286354} and \url{https://github.com/visionlab-ogu/PLANesT-3D/tree/main/data}}) introduced in this work contributes to the diversity of publicly available annotated 3D plant models in terms of acquisition modality, number of distinct plants, and inclusion of other species. The data is acquired with an RGB camera and reconstructed through structure from motion (SfM) and  multi-view steoroscopy (MVS). Together with the Soybean-MVS dataset, the PLANesT-3D dataset will balance the dominant acquisition modality, which is laser scanning. Reconstruction with SfM and MVS does not require expensive equipment or elaborate acquisition setups, while yielding both geometric and textural information. 

While Pheno4D and Soybean-MVS datasets are extremely valuable since they track the development of individual plants across time, the number of distinct plants  are 14 for Pheno4D and 5 for Soybean-MVS. The PLANesT-3D dataset includes models of 34 distinct plants belonging to three species: namely \textit{Capsicum annuum} (pepper), \textit{Rosa kordana} (rose), and \textit{Ribes rubrum} (ribes). Also, point clouds from two species (pepper and ribes), 3D models of which were not available before, are added to the public domain. The diversity of training and evaluation data in terms of acquisition modality, noise levels, plant species, plant instances, plant architecture, organ geometry, etc. plays an important role in assessing the robustness as well as the generalization ability of machine learning approaches.

The objective of this study is twofold: First is to introduce PLANesT-3D, a new dataset containing complete 3D color point clouds of plants together with their semantic and instance labels. Second is to present a new segmentation approach, the SP-LSCnet\footnote{The code for SP-LSCnet is available at \url{https://github.com/visionlab-ogu/PLANesT-3D}}, that combines an unsupervised clustering scheme and an adaptive network for point cloud classification. 

As the capacity of machine learning techniques to model a large range of variations in data increases rapidly, more and more training samples are required to exploit the potential of those techniques. In the domain of 3D plant phenotyping, both the diversity and amount of labeled data acquired from real plants are very important. Even for a single plant species, more data is required for machine learning techniques to model the variations among plant instances in terms of structural and geometric complexity. PLANesT-3D introduces new instances from three different species, with a large within-species variety in terms of plant size, architectural complexity, and the number and density of organs, such as leaves.

Introducing annotated 3D models of different species is also key to train machine learning techniques and compare their merits and shortcomings over multiple species. In this respect, PLANesT-3D contributes to the field with annotated models of two new species, pepper and ribes. These species, especially ribes, bring in more variability in terms of size, architecture, and geometric variations of the organs. The ribes plants in our dataset vary in size from 30 cm to 1m. The leaves are many in quantity (reaching 100 for some individual plants) and dense compared to plants in existing datasets. They also possess distinct geometric structures. The diversity in terms of the shapes of the leaves and stems of pepper, rose and ribes plants enriches the collection of available datasets.

Provision of annotated 3D models of various species is indispensable to assess the generalization ability and transferability of a machine learning model to instances of a different species. Also, such trained models can be re-trained for comparison of various domain-adaptation techniques with respect to the amount of new annotated data they require to effectively process instances of a new species. Although not explored in this study experimentally, we envision that PLANesT-3D will enable researchers to test their machine learning models on different species, instances of which were not present in their training sets.

Plant phenotyping and plant monitoring systems operate with acquisition devices ranging from single cameras and multi-camera setups to RGB-D sensors and laser scanners. The nature of the acquired 3D data differs widely depending on the sensor in terms of point density, and the structure and amount of noise. While simulating noise and varying point density is possible by virtual manipulation of existing point clouds, 3D models originating from various acquisition systems are precious to assess the robustness of phenotyping systems. As distinct from existing datasest, PLANesT-3D dataset introduces 3D plant models built from images acquired by a single hand-held camera. This particular reconstruction technique resulted in point clouds which contain highly-dense and high-quality regions together with sparse and noisy regions due to heavy self-occlusion. 

The second objective of this study is to introduce a new semantic segmentation approach, where the 3D data points are mapped to 2D domain for clustering and the corresponding 3D point cloud of each cluster is processed separately with a classification network. State-of-the-art deep learning methods processing plant point clouds are mostly designed as end-to-end networks, whose layers and modules are tuned to the training data. Also, such systems operate as a black-box and do not offer much interpretability to the end-user.

We would like to showcase an alternative to this approach, the SuperPoint-Leaf-Stem-Classification network (SP-LSCnet), where a portion of the system is unsupervised, easy to visualize, and interpretable. The unsupervised clustering scheme maps the 3D point cloud into the 2D space via t-distributed stochastic neighborhood embedding (t-SNE), which clusters points on the same manifold together and preserves the local structure of the clusters. This behavior of t-SNE is well-suited for 3D plant data, which, when mapped to 2D, looks like flattened versions of parts of 3D leaves and branches. The subsequent unsupervised clustering method allows the user to visualize the clusters in a single map, where the nature of each cluster is visible. The method does not necessitate training, and is readily applicable to new plant species.  

The subsequent classification stage in the new SP-LSCnet method operates on the 3D counterparts of the clusters obtained in 2D. Each cluster is processed as a separate point cloud and processed by an adaptive network for classification into a leaf or a stem. The advantage of this modular approach is that the clusters can be processed by any other classification network, allowing the user to track and compare the results visually on the 2D map of clusters. It is also possible to convert this visualization ability into an interface where the user can correct wrong predictions. Such an interface can be a basis of a fast annotation tool, where the annotator can observe a complex 3D plant point cloud through a 2D mapping of organs and parts of organs with class labels predicted by a network and can correct classification errors quickly.

In summary, PLANesT-3D is provided to help close the gap between data scarcity in terms of amount and diversity of 3D plant models and the progress on the machine learning methods for segmentation of 3D plant models. PLANesT-3D introduces annotated point clouds of plants of two new species, which were not present in the existing datasets. Also, PLANesT-3D, as opposed to the existing datasets, provides point clouds that were acquired with a low-cost modality that do not require specialized setups. In this way, PLANesT-3D will contribute to the diversity of available data for assessing generalization ability of machine learning methods for 3D plant phenotyping. We also introduce the SP-LSCnet method to demonstrate that a partly-unsupervised segmentation scheme, with intermediate 2D visualization options yields competitive performance. These intermediate 2D visualizations greatly increase the interpretability of our method in contrast to black-box deep neural networks. Its unsupervised nature and modularity increases its ability to generalize to new species without extensive training. In addition, we provide semantic segmentation results on the PLANesT-3D dataset with two point-based deep learning architectures, PointNet++ and RoseSegNet. 

The main contributions of this work can be summarized as follows:

\begin{enumerate}
\item We introduce PLANesT-3D, a new dataset containing RGB point clouds representing 34 real plants from three species.
\item We propose SP-LSCnet as a new semantic segmentation approach where an unsupervised clustering method based on t-SNE is combined with a point cloud classifier network. The classifier network employs two adaptive modules to adjust local region organization for feature extraction. 
\item We tested RoseSegNet, a semantic segmentation algorithm we had previously developed for plant data \citep{Turgut2022rosesegnet} on the new dataset, and demonstrated that it is effective without requiring hyperparameter re-adjustment.

\end{enumerate}

The rest of this article is organized as follows. In Sections \ref{sec:dataAcq} and \ref{sec:reconst}, we describe the data acquisition, reconstruction, and pre-processing steps leading to the creation of 3D point clouds. We give detailed information about the PLANesT-3D dataset in Section \ref{sec:dataset}. The segmentation methods evaluated in this work are described in Section \ref{sec:threemethods}. Our new segmentation approach, SP-LSCnet, is introduced in Section \ref{sec:SPLSCnet}. The performances of the semantic segmentation methods on the PLANesT-3D dataset are reported in Section \ref{sec:results}. A discussion of the work is provided in Section \ref{sec:discussion}, followed by our conclusion in Section \ref{sec:conclusions}.

\begin{figure*}[!ht]
\centering
\includegraphics[width=0.75\textwidth]{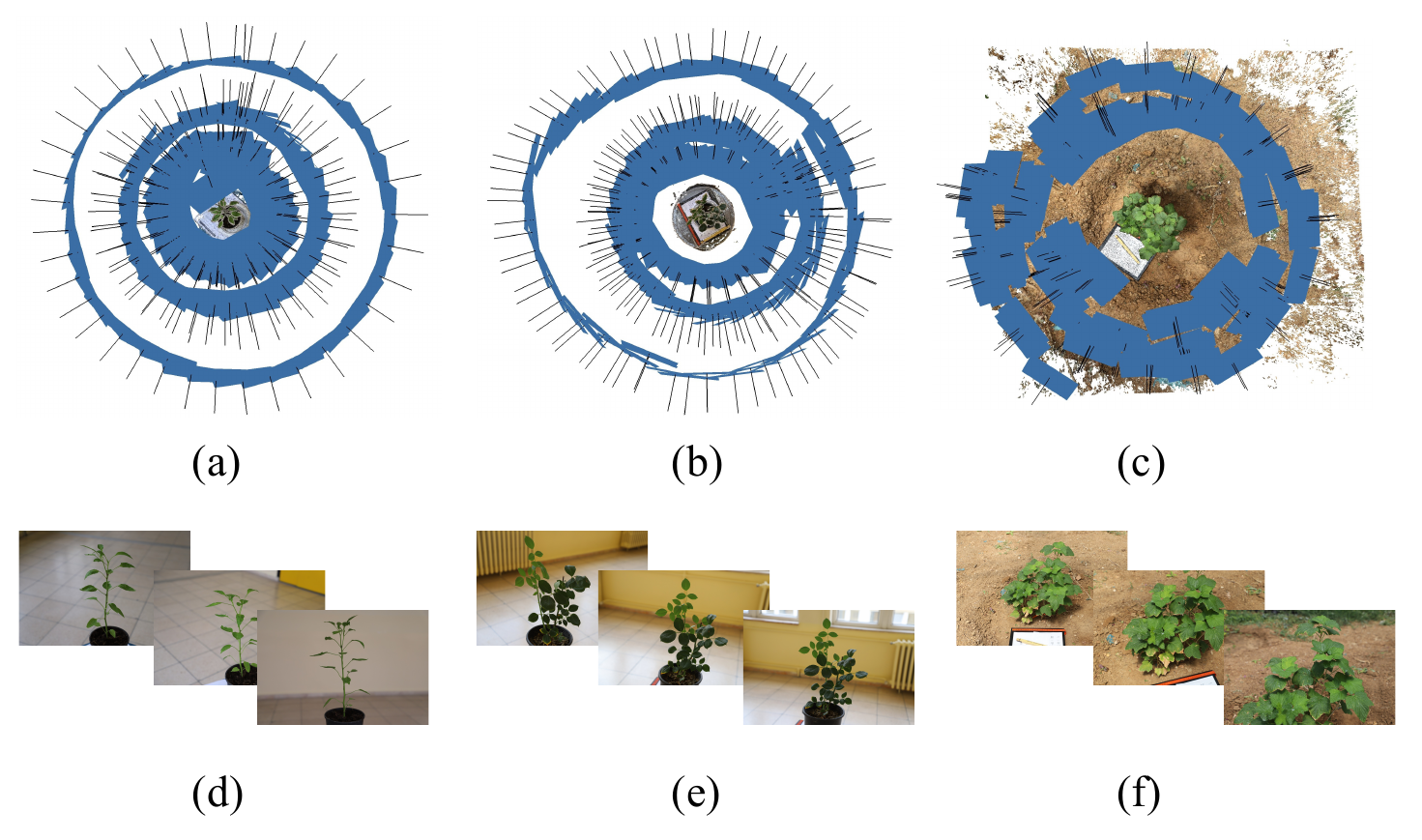}
\caption{Estimated camera poses for 231 images of a pepper plant (a), 240 images of a rose plant (b), and 177 images of two ribes plants (c). Sample images for pepper (d), rose (e), and ribes (f) are also provided.}
\label{fig:cameraPose}
\end{figure*}

\section{Materials and Methods}
\label{sec:methods}

\subsection{Data acquisition}
\label{sec:dataAcq}

Plants have intricate architectures and individuals of a single species possess significant diversities in terms of structure, shape and texture. The number of individual plants and the representativeness of the variations in a training dataset are essential for machine learning algorithms to model a specific species effectively. However, generating manually annotated data on dense and complicated point clouds is time intensive and the number of models to be annotated is restricted by available human resources.

We targeted to reconstruct and annotate 10 individual plants for each of the three species for our new dataset: pepper (\textit{Capsicum annuum}), rose (\textit{Rosa kordana}), and ribes  (\textit{Ribes rubrum}), which are considered to be valuable produce. Rose models were already available in the ROSE-X dataset \cite{Dutagaci2020rose}, however they were acquired with X-ray imaging. The new rose models in our dataset possess reconstruction noise and variations of point density. Models of the other two species were not present in other datasets. While selecting individual plants, we considered variations in terms of plant size, general shape and architecture, number of leaves, and the shape and texture of the branches and leaves. As a result of this selection process, we modeled 10 potted pepper plants, 10 potted rose plants, and 14 ribes plants planted directly to the soil in a greenhouse.

In order to obtain full 3D plant models, 2D color images of 34 plants were acquired using a handheld DSLR camera (EOS 6D MARK II, Canon, Japan). Multiple color images of resolution $6240 \times 4160$  were captured manually around each plant via positioning the camera so as to cover as much plant surface as possible with a high degree of overlap. Three examples of the sets of camera locations and orientations are illustrated in Fig. \ref{fig:cameraPose}.

Table \ref{tab:numimages} gives the number of images used to reconstruct the 3D plant point clouds. Three pairs of Ribes were very closely planted (Ribes 02 \& 12, Ribes 04 \& 13, Ribes 05 \& 14). A single point cloud was reconstructed for each pair using the set of images where the pair appeared together. After reconstruction, and pose normalization, the two plants were manually segmented into separate point clouds of two individual plants. 

\begin{table}[!ht]
  \centering
  \caption{Number of images used to reconstruct the 3D plant point clouds.}
  \label{tab:numimages}
  \scriptsize
  \begin{tabular}{lc|lc|lc}
    \hline
    Plant ID & \# images & Plant ID & \# images  & Plant ID & \# images   \\
    \hline
    Pepper 01 &  261 &  Rose 01 & 210 & Ribes 01 & 243\\
    \hline
    Pepper 02 &  320 &  Rose 02 &  182 & Ribes 02 \& 12 & 177\\
    \hline
    Pepper 03 &  324 &  Rose 03 & 203 & Ribes 03 & 294\\
    \hline
    Pepper 04 & 282  &  Rose 04 & 205 & Ribes 04 \& 13 & 235\\
    \hline
    Pepper 05 & 280 &  Rose 05 &  240 & Ribes 05 \& 14 & 169\\
    \hline
    Pepper 06 & 231  &  Rose 06 &  211 & Ribes 06 & 150\\
    \hline
    Pepper 07 & 294 &  Rose 07 & 258 & Ribes 07 & 225\\
    \hline
    Pepper 08 & 350 &  Rose 08 & 406 & Ribes 08 & 222\\
    \hline
    Pepper 09 & 199  &  Rose 09 & 377 & Ribes 09 & 191\\
    \hline
    Pepper 10 & 226  &  Rose 10 & 386 & Ribes 10 & 310\\
    \hline
     &   &  &  & Ribes 11 & 240\\
    \hline
  \end{tabular}
\end{table}

\subsection{Reconstruction and pre-processing of 3D point clouds}
\label{sec:reconst}

\begin{figure*}[!t]
\centering
\includegraphics[width=0.4\textwidth]{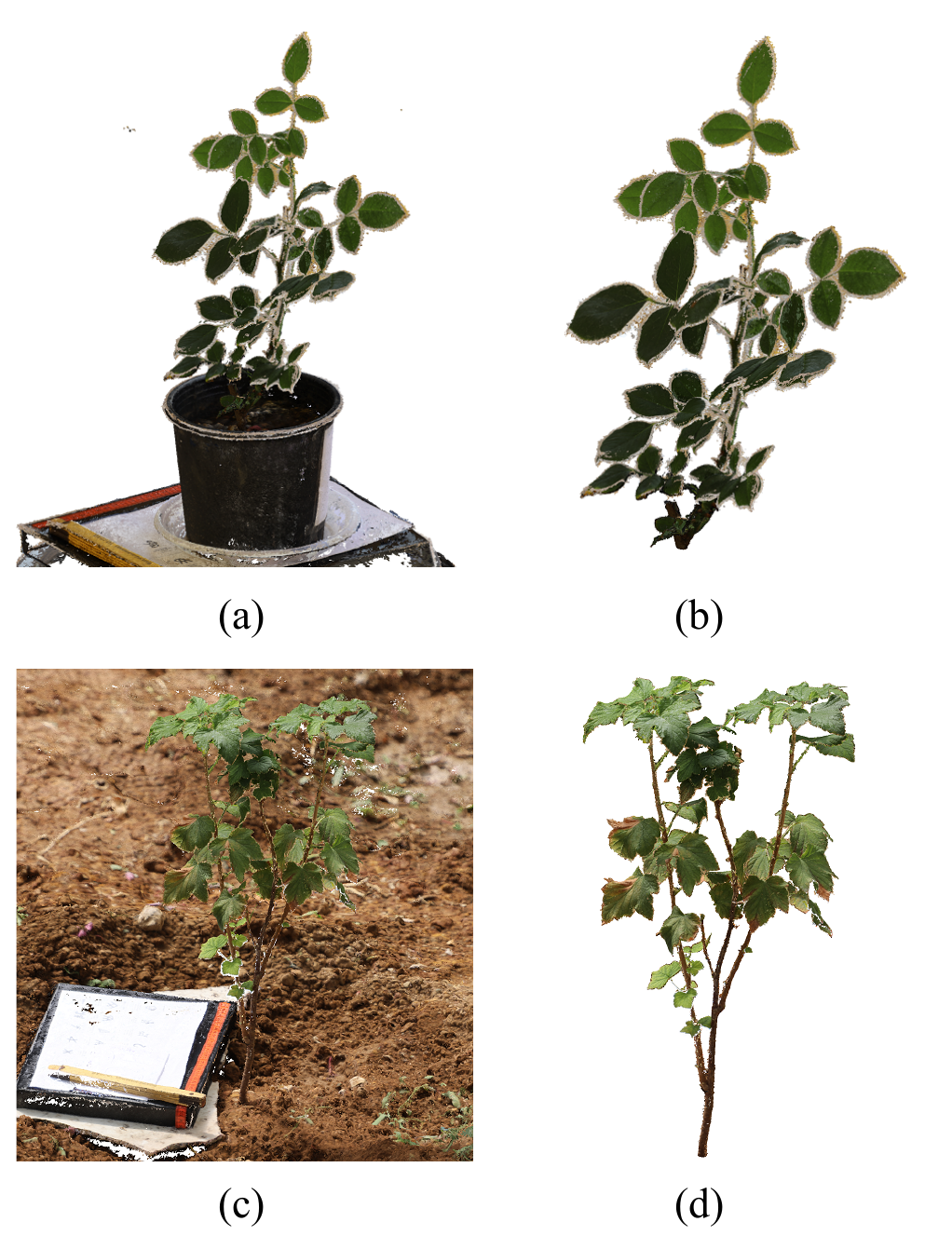}
\caption{Raw point clouds reconstructed via Agisoft Metashape Professional of a rose plant (a) and a ribes plant (c). Corresponding clouds including only 3D points of the target plants are given in (b) for rose, and in (d) for ribes.}
\label{fig:agisoftROI}
\end{figure*}

For reconstruction of 3D color point clouds from 2D color images, Agisoft Metashape Professional (Agisoft LLC, St. Petersburg, Russia) was employed. Agisoft Metashape is a software that performs photogrammetric processing of digital images and generates 3D models through principles of structure from motion (SFM) and Multi-view Stereo (MVS) \citep{hartley2003multiple}. Examples of the raw point clouds produced by Metashape Professional are given in Fig. \ref{fig:agisoftROI}. The software also provides confidence maps for the 3D points. The confidence value for a point corresponds to the number of images that "see" and contribute to reconstruct the point. 

As can be observed from Fig. \ref{fig:agisoftROI}, the raw point clouds include structures from the background and are noisy. The correct scale of the scene is lost. Automatic segmentation of plants from each other and from the background is in itself a research problem. However, we aimed to provide a dataset of 3D models that correspond to isolated plants for evaluation of algorithms that focus on the analysis of single plants.

In order to recover the correct scale of the point cloud, remove background structures and noise, and segment the plant points from the scene, we devised a simple, semi-automatic procedure. The manual operations involve the following: 1) Picking landmark points from the raw 3D point cloud to recover the scale. 2) Measuring the distances between landmark points in the real scene. 3) Picking a point on the plant close to the plant base. The first two operations are used to scale the point cloud with the scale factor, which is the ratio of the measurements on the point cloud and on the real scene. The recovery of scale through landmark points is a common practice for 3D reconstruction with SfM. The third operation is used to mark the center of the new coordinate frame. Apart from these three manual interventions, the rest of the procedure is performed automatically.

We denote the raw point cloud as $\mathcal{P}_R$, where a point $p \in \mathcal{P}_R$ is represented by its coordinates $p = (x, y, z)$ defined in the coordinate frame returned by Metashape. The steps of the procedure are detailed as follows: 

\begin{itemize}

\item \textbf{Recovering the scale:} 
The multi-view reconstruction pipeline employed by Metashape Professional provides the scene geometry up to a scale, meaning that the correct scale of the structure is lost. To enable recovery of the correct scale, we installed rulers and/or objects with distinct patterns in the vicinity of the plants during image acquisition. Once the 3D point cloud is reconstructed, a number of landmarks such as the markers on the rulers and/or on the patterns were picked manually. Let the set of these landmark points be $\{p_r\} \subset \mathcal{P}_R$ with $r=1,2,...,N_r$, where $N_r$ is the number of landmark points. The distance between a pair of these reference points on the point cloud was determined as $\hat{d}_{rs} = \| p_r - p_s\|$, where $\| \cdot \|$ corresponds to the Euclidean norm. The average of the distances of all the pairs is equal to
\begin{equation}
\hat{D}_R = \frac{1}{N_r(N_r-1)} \sum_{r \neq s}{\sum_{s}{\hat{d}_{rs}}}\;.
\end{equation}

Let $D_{R}$ be calculated using the true metric distances (in cm units) between the landmark points $d_{rs}$ measured manually, or read through the ruler. The scale factor is then determined as 

\begin{equation}
S = \frac{D_{R}}{\hat{D}_R}\;.
\end{equation}

The point cloud is scaled such that the average of the reference distances on the point cloud is equal to the measured distances on the real objects, i.e. the point coordinates were updated as $p'= S \cdot p$ for all $p \in \mathcal{P}_R $. The scaled point set is denoted as  $\mathcal{P}'_R$.

\item \textbf{Rotating the point cloud to a normalized pose:} We aimed to rotate the reference frame such that the XY-plane corresponded to the ground or the plane holding the plant pot. In order to estimate the parameters of this plane, we used M-estimator SAmple Consensus (MSAC) algorithm given in \citep{Torr2000}, which is a variant of RANdom SAmple Consensus (RANSAC) algorithm. We manually picked a point, $p'_{base}$, close to the plant base. Using the estimated normal vector $\mathbf{n} = [ n_x,n_y,n_z ] $ of the plane in the scene and the point $p'_{base}$ at the plant base, we translated and rotated the point cloud $\mathcal{P}'_R $ into a new reference frame using

\begin{equation}
\hat{p} = \mathbf{R}(p'-p'_{base})
\label{eq:rot}
\end{equation}

for all $p' \in \mathcal{P}'_R$. The point cloud rotated to this normalized pose is denoted as  $\hat{\mathcal{P}}_R$. In Eq. \ref{eq:rot}, $\mathbf{R}$ corresponds to the rotation matrix which transforms the normal $\mathbf{n}$ to $[0, 0, 1]$ in the new reference frame. The origin of this reference frame corresponds to the hand-picked point at the plant base. The XY plane coincides with the plane detected by the MSAC algorithm. The positive Z-direction is oriented from the plant base towards to the shoot. 

\item \textbf{Extracting plant points:}
Once the origin of the reference coordinate frame is established at the plant base, and the positive Z-direction points toward the plant shoot, we removed all points with a negative $z$ coordinate, assuming all plant points remain above the plant base. We verified this assumption for all the models in the dataset. We applied connected component analysis to the rest of the point cloud and retained the points belonging to the largest connected component as the final plant point set $\mathcal{P} \subset \hat{\mathcal{P}}_R$. Examples of the outputs of this semi-automatic process is given in Fig. \ref{fig:agisoftROI}b for a rose plant and in Fig. \ref{fig:agisoftROI}d for a ribes plant. 
\end{itemize}

In summary, the automatic operations in this semi-automatic method involve 1) detecting a plane through MSAC to establish the orientation of the XY plane of the reference frame; 2) using the manually picked point close to the plant base to direct the positive direction of Z axis towards the plane shoot, and translate the origin of the reference frame; 3) removing points below the XY plane; and 4) applying connected components analysis to select the largest component as the plant point cloud. One assumption for this procedure to work for other settings of 3D plant data acquisition is the availability of a prominent plane (the ground or a platform) that is relatively perpendicular to the plant orientation. Another assumption is that the plant parts should not trail below the supporting surface.

\begin{figure*}[!ht]
\centering
\includegraphics[width=0.6\textwidth]{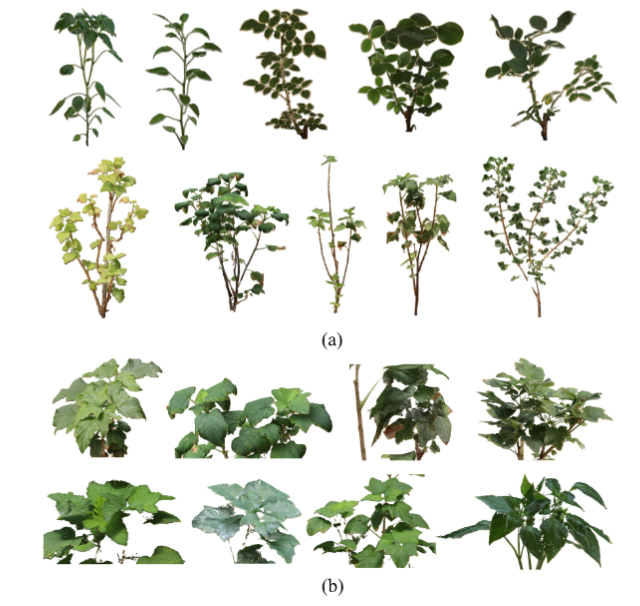}
\caption{Sample point clouds from PLANesT-3D demonstrating the diversity in terms of the overall plant shape and architecture (a).  Examples of dense foliage in the PLANesT-3D dataset (b).}
\label{fig:overallShapeFoliage}
\end{figure*}

\begin{figure*}[!ht]
\centering
\includegraphics[width=0.6\textwidth]{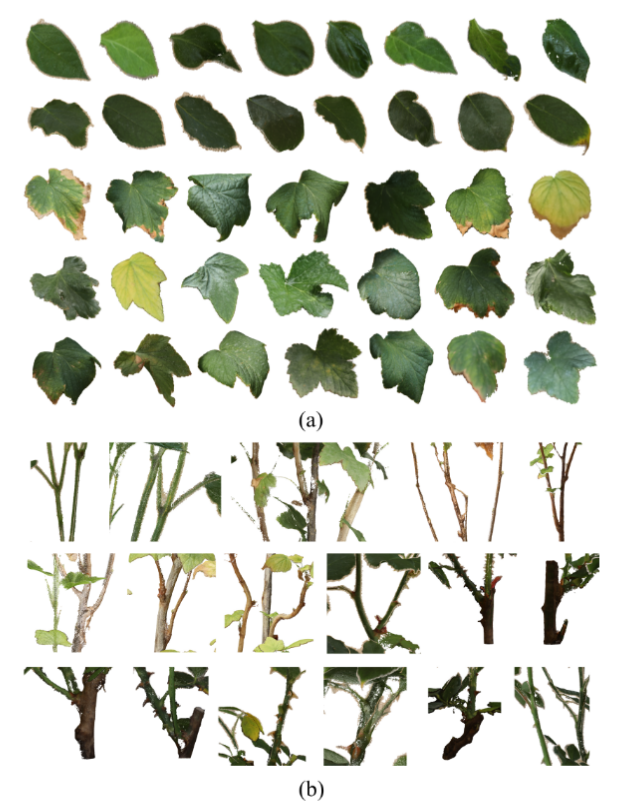}
\caption{Diversity in leaf shape and texture (a). Diversity in stem shape (b).} 
\label{fig:leaves_branches}
\end{figure*}

\begin{figure*}[!ht]
\centering
\includegraphics[width=0.7\textwidth]{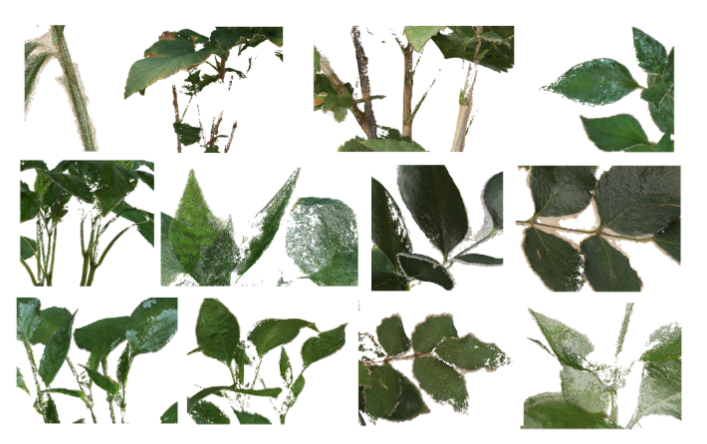}
\caption{Instances of noisy and missing data due to self-occlusion and reconstruction errors.}
\label{fig:noise}
\end{figure*}

\subsection{The PLANesT-3D dataset}
\label{sec:dataset}

The PLANesT-3D dataset consists of 34 plant point clouds, where the plant points are isolated from the background as described in the previous section. Instances of three species are present in the dataset. The global shape and architecture, especially for the Rose and Ribes plants are diverse in PLANesT-3D as demonstrated in Fig. \ref{fig:overallShapeFoliage}a. The number of leaves vary between 21 and 105, and the height of the plants is in the relatively large range of 25 to 105cm. The density of the foliage varies from plant to plant, with some plants having high-density foliage as shown in Fig. \ref{fig:overallShapeFoliage}b. The closely packed leaves pose a particular challenge for leaf instance segmentation methods.

The PLANesT-3D dataset is rich in the geometric and textural variations of leaf and stem instances. Fig. \ref{fig:leaves_branches}a shows isolated leaves of high point density from the dataset. The leaves of Ribes plants are of considerable diversity, both in shape and appearance. Stems of Ribes and Rose plants have diverse shape characteristics as illustrated in Fig. \ref{fig:leaves_branches}b, such as the presence of cut stems, varying stem thickness and curvature, varying branching angles, and variety in the arrangements of thorns.

We selected high point density leaves for illustrating the shape variations in Fig. \ref{fig:leaves_branches}a. However, due to the presence of thin leaves and branches and heavy self-occlusion around high density foliage, the point clouds have structures with missing data and reconstruction noise. Instances of such cases are given in Fig. \ref{fig:noise}. Noise is especially present at boundaries with color bleeding from the background. Color bleeding is the phenomena where colors from one area of the scene spreads into other objects, especially into the borders. There is also reconstruction noise present on the surfaces of leaves and stems. Parts of heavily occluded branches and leaves were not observed from any of the camera positions, leading to missing portions in the 3D models. Some small leaves and thin branches were not fully reconstructed resulting from the limited resolution of the 2D images.

\begin{figure*}[!ht]
\centering
\includegraphics[width=0.6\textwidth]{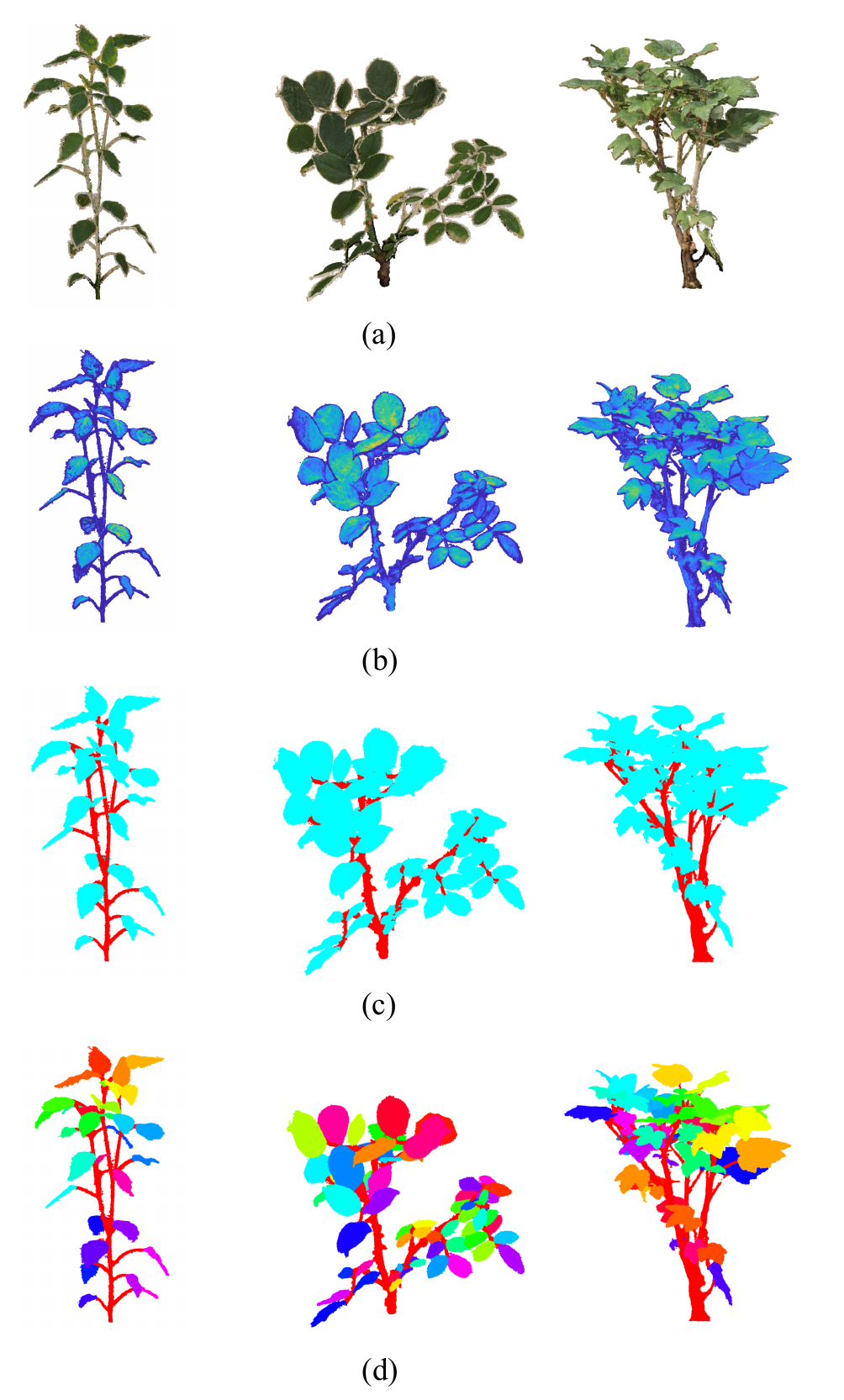}
\caption{Three models from the PLANesT-3D dataset rendered with plant color (a), confidence map (b), semantic labels (c), and instance labels (d).}
\label{fig:modelSamples}
\end{figure*}

The point clouds in the PLANesT-3D data possess color information as reconstructed by Metashape Professional. One important fact about the acquisition of the 2D images is that the illumination was not controlled. While all the camera images for an individual plant were acquired in one single session, image data from different plants were collected at different days and times of days. The illumination in the environment was not controlled. That is apparent from Fig. \ref{fig:leaves_branches}a, where apart from the natural variations in leaf texture, part of the color diversity comes from illumination changes, such as highlights and shadows due to the direction and amount of light.

As a result of these observations, we can conclude that the point clouds in PLANesT-3D dataset include leaf and stem instances of both high and low point density and quality with varying shape, texture and color characteristics. These  characteristics make the PLANesT-3D a challenging dataset for modeling various variations in realistic data acquisition scenarios.

Both semantic labels and instance labels of the 34 plant point clouds were obtained through manually labeling each point using CloudCompare point cloud processing software \citep{cloudcompare}. Semantic labels for PLANesT-3D correspond to "leaf" and "stem" classes. Points on the leaf blades were labeled as "leaf"; main stem, branches, and petioles were included in the "stem" class. Instance labels are identity numbers representing individual leaflets within a plant model. A single instance label is assigned to all points semantically labeled as "stem".

Table \ref{tab:properties} gives five properties of each of the 34 plant point clouds in the dataset: 1) Plant height, 2) Total number of points, 3) Percentage of "leaf" points, 4) Percentage of "stem" points, and 5) Number of individual leaflets. Plant height is in the range of 25 to 105cm. The number of points exceed 1 million for most plants. Number of individual leaflets varies between 21 to 105, indicating the variation of complexity in plant structure.

Accompanying the locations and color information of the points in each model, three scalar fields are provided: The confidence map as delivered by Metashape Professional, semantic labels, and instance labels. Examples of point clouds from the PLANesT-3D dataset and their corresponding confidence maps and labels are depicted in Fig. \ref{fig:modelSamples}.

\begin{table*}[!hb]
  \centering
  \caption{Properties of the point clouds in PLANesT-3D}
  \label{tab:properties}
  \begin{tabular}{lccccc}
    \hline
     & Height (cm) & \# points & "leaf" (\%)  & "stem" (\%) & \# leaflets \\
      
    \hline
    Pepper 01 &  53.63 &  2312959 & 70.45 & 29.56 & 33\\
    \hline
    Pepper 02 &  63.39 &  5531852 &  76.80 & 23.20  & 48\\
    \hline
    Pepper 03 &  60.65 &  3482959 & 72.43 & 27.57 & 45\\
    \hline
    Pepper 04 & 51.51  &  2425889 & 68.28 & 31.72 & 40\\
    \hline
    Pepper 05 & 48.95 &  2720311 &  69.52 & 30.48 & 35\\
    \hline
    Pepper 06 & 32.94  &  1846694 &  65.41 & 34.59 & 21\\
    \hline
    Pepper 07 & 49.71 &  1806010 & 69.11 & 30.89 & 27\\
    \hline
    Pepper 08 & 48.45 &  2311602 & 69.46 & 30.54 & 27\\
    \hline
    Pepper 09 & 53.32  &  3440529 & 74.10 & 25.90 & 38\\
    \hline
    Pepper 10 & 64.95  &  2416062 & 69.91 & 30.09 & 36\\
    \hline
    Rose 01 &  29.96 &  3452545 & 80.15 & 19.85 & 59\\
    \hline
    Rose 02 &  33.55 &  3601210 & 80.55 & 19.45 & 68\\
    \hline
    Rose 03 &  32.96 &  2580315 & 75.97 & 24.03 & 43\\
    \hline
    Rose 04 & 27.36  &  4091952 & 82.37 & 17.63 & 61\\
    \hline
    Rose 05 & 38.01 &  2867159 &  78.29 & 21.71 & 65\\
    \hline
    Rose 06 & 25.98  &  2540487 &  81.62 & 18.38 & 27\\
    \hline
    Rose 07 & 27.88 &  2999376 & 82.24 & 17.76 & 43\\
    \hline
    Rose 08 & 46.80 &  3616281 & 74.12 & 25.88 & 100\\
    \hline
    Rose 09 & 50.16  &  2119400 & 77.45 & 22.55 & 56\\
    \hline
    Rose 10 & 37.65  &  3296993 & 77.65 & 22.35 & 85\\
    \hline
    Ribes 01 &  76.92 &  2078729 & 88.33 & 11.67 & 64\\
    \hline
    Ribes 02 &  33.69 &  1299910 & 89.67 & 10.33 & 34\\
    \hline
    Ribes 03 &  74.25 &  1053341 & 84.32 & 15.68 & 33\\
    \hline
    Ribes 04 & 34.64  &  1190366 & 85.32 & 14.68 & 32\\
    \hline
    Ribes 05 & 104.85 &  1506387 & 78.09 & 21.91 & 105\\
    \hline
    Ribes 06 & 31.55  &  3175798 &  81.49 & 18.51 & 63\\
    \hline
    Ribes 07 & 93.66 &  3553784 & 85.66 & 14.34 & 101\\
    \hline
    Ribes 08 & 79.73 &  600792 & 67.04 & 32.96 & 41\\
    \hline
    Ribes 09 & 43.53  &  679885 & 68.20 & 31.80 & 34\\
    \hline
    Ribes 10 & 62.12  &  1256866 & 82.29 & 17.71 & 54\\
   \hline
    Ribes 11 & 33.32 & 1194023 &   61.12 & 38.88 & 99\\
    \hline
    Ribes 12 & 52.52  &  3147426 & 87.08 & 12.92 & 70\\
    \hline
    Ribes 13 & 46.18  &  3258674 & 79.56 & 20.44 & 92\\
   \hline
   Ribes 14 & 56.13  &  644626 & 83.33 & 16.67 & 21\\
   \hline
  \end{tabular}
\end{table*}

\begin{figure*}[!ht]
\centering
\includegraphics[width=0.5\textwidth]{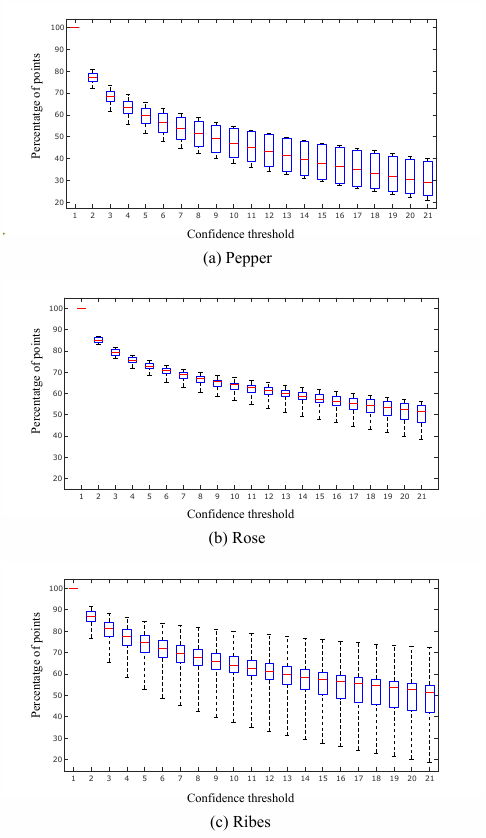}
\caption{Percentage of remaining points as the confidence threshold varies.}
\label{fig:confidencePERC}
\end{figure*}

\begin{figure*}[!ht]
\centering
\includegraphics[width=0.95\textwidth]{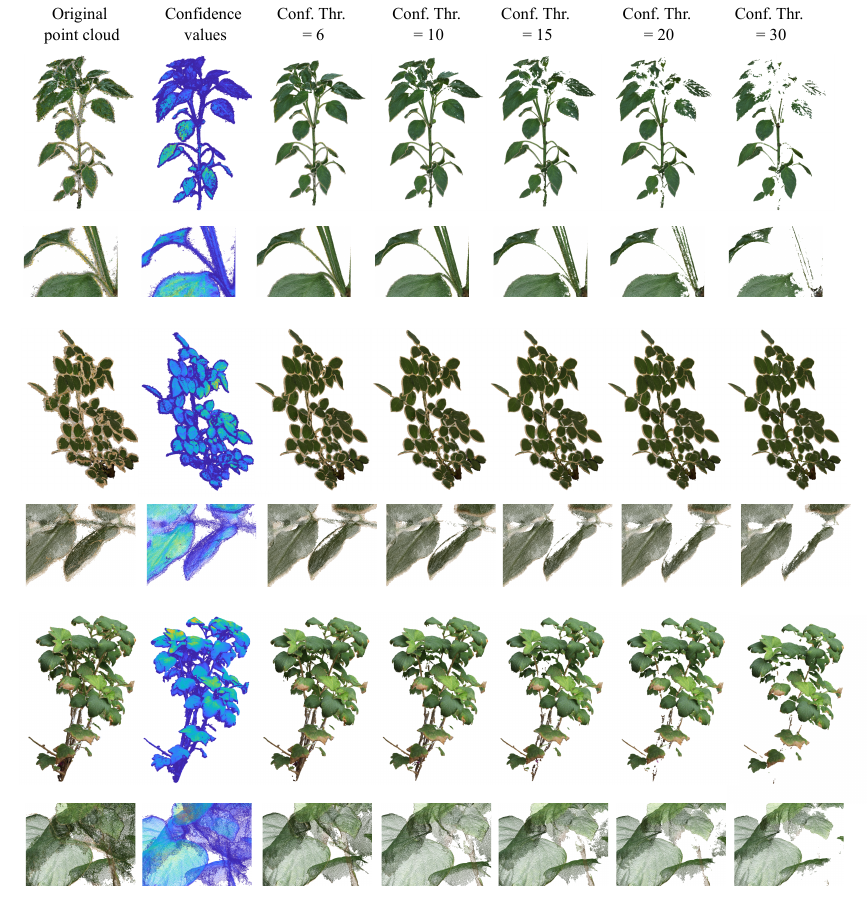}
\caption{The effect of confidence threshold.}
\label{fig:confidenceVisuals}
\end{figure*}

\begin{figure*}[!ht]
\centering
\includegraphics[width=0.5\textwidth]{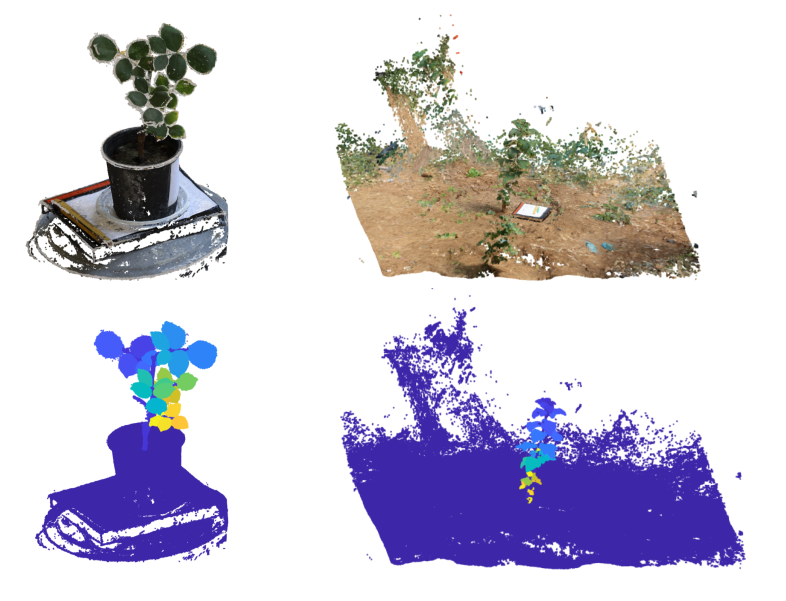}
\caption{Examples of raw point clouds with instance labels of the target plant.}
\label{fig:origRaws}
\end{figure*}

As stated before, the confidence value for a point corresponds to the number of images where the point is observable, and that contribute to the 3D reconstruction of the point. It can be viewed as a measure of confidence on the accuracy of the estimated 3D location. The point cloud can be filtered by discarding points that have confidence values less than a certain threshold. Fig. \ref{fig:confidencePERC} gives the box plots of the percentage of the number of points in a 3D model with confidence values above and equal to the confidence thresholds varying between 1 and 21. A separate box plot is given for each species. The bottom and top of each box are the 25th and 75th percentiles among the point clouds in the species, the whiskers indicate the range and the red line in the middle of each box gives the median value. We can observe a significant drop in the number of points as the threshold for the number of cameras observing the point increases. The drop in percentage is lower and less varied for the rose plants since the rose plants are smaller in size making it easier to cover the plant surface with more camera viewpoints observing smaller areas. More importantly, the rose plants suffer from self-occlusion to a lower degree compared to Pepper and Ribes.

Fig. \ref{fig:confidencePERC} demonstrates the effect of filtering point clouds with varying confidence thresholds. As the number of camera viewpoints required to reconstruct a point increases, the highly visible parts remain relatively of high point density. The quality of broad leaves which are oriented perpendicularly to camera view orientations is not degraded. Rather, reconstruction noise on the surface of such leaves is reduced. However, around dense foliage, such as the top of the pepper and ribes plants, small leaves, and leaves oriented in parallel to the camera view orientations loose significant amount of points. Also, thin branches and petioles and branches and leaves hidden by heavy self-occlusion begin to disappear. The confidence values are provided as a scalar field of the models in the PLANesT-3D dataset for further investigation of the trade-off between reduced reconstruction noise and loss of self-occluded and fine-detail data.

In addition to the labeled 3D point clouds of individual plants, which were isolated from the background, we also provide the full raw point cloud as reconstructed by Metashape Professional. The instance labels were mapped onto the plant points in the scene. Examples of such point clouds are given in Fig. \ref{fig:origRaws}. These point clouds will be instrumental to test algorithms that automatically separate target plants from the environment. The 2D color images for all the 34 plants together with their estimated camera poses and parameters are also open to the public to provide input data for recent 3D reconstruction techniques\footnote{The data is available at \url{https://github.com/visionlab-ogu/PLANesT-3D/tree/main/data}}.

\subsection{Methods for semantic segmentation}
\label{sec:threemethods}

\citet{harandi2023make} provided a comprehensive review of typical steps involved in the processing and analysis of 3D representations of plants, including point clouds. Models in PLANesT-3D can be used to evaluate a diverse set of tools developed for plant analysis and phenotyping. In this study, we focus on the application of segmentation of 3D point clouds of plants into their semantic units.

The objective of semantic segmentation is to assign each point $p$ in the point set $\mathcal{P}$ one of the semantic labels. The semantic categories in the PLANesT-3D dataset correspond to "leaf" and "stem", where "leaf" points are on leaf blades, and "stem" points belong to the main stem, branches, and petioles.

We developed a novel semantic segmentation method that is a combination of a superpoint extraction scheme \citep{dutagaci2023using} and an adaptive 3D object classification network \citep{turgut2023local}. We abbreviate this method as SP-LSCnet, where SP stands for superpoint, while LSCnet is a leaf-stem classification network. In order to provide reference performance results for further research, we also tested PointNet++ \citep{qi2017pointnet2} and RoseSegNet \citep{Turgut2022rosesegnet} on the new PLANesT-3D  dataset.

\subsubsection{SP-LSCnet}
\label{sec:SPLSCnet}

SP-LSCnet consists of two stages: Superpoint extraction and leaf-stem classification. In superpoint extraction stage, the point cloud is over-segmented into superpoints through an unsupervised process that operates on 2D points embedded by t-SNE. Then, each superpoint generated in the first stage is classified through a deep neural network equipped with attention-based modules that adaptively determine receptive fields for feature extraction. The flowchart of the method is depicted in Fig. \ref{fig:flowSeg}, and the details of the two stages are given below:

\begin{figure*}[!ht]
\centering
\includegraphics[width=0.98\textwidth]{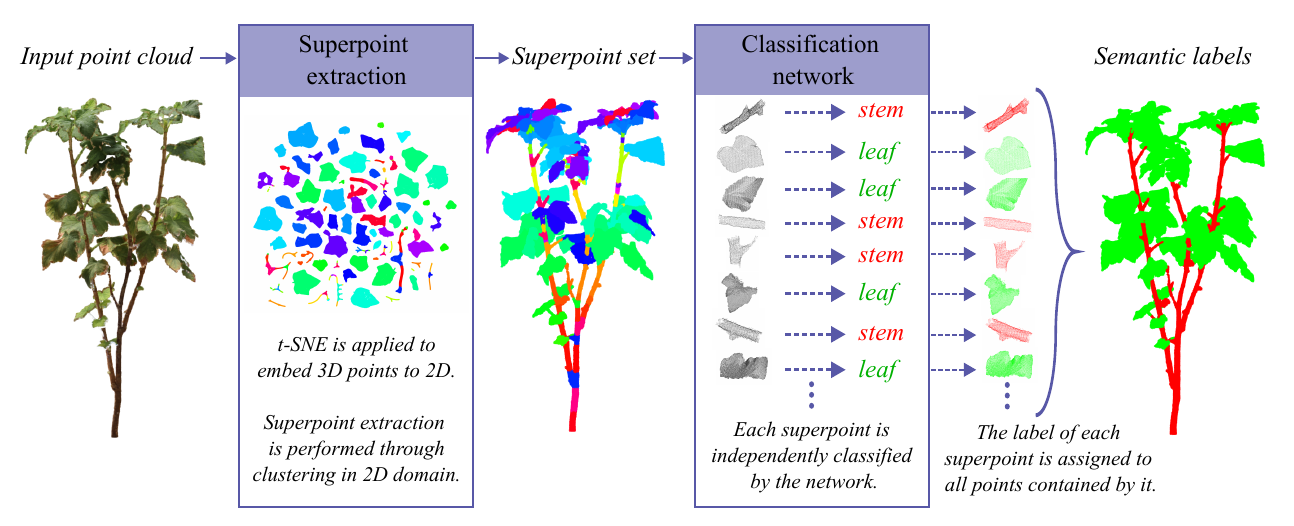}
\caption{Flowchart of SP-LSCnet. The input point cloud is first passed through the superpoint extraction procedure. This procedure involves embedding of the 3D points into 2D and partitioning the 2D points into superpoints. Once the superpoint assignments are mapped to 3D domain, each superpoint is passed independently through a leaf-stem classification network. The 3D points in a superpoint are all labeled with the global class of the superpoint predicted by the network.}
\label{fig:flowSeg}
\end{figure*}

\underline{\emph{Superpoint extraction:}} For accomplishing this task, the superpoint extraction scheme proposed in \citep{dutagaci2023using} is selected. The plant point cloud is first downsampled with voxel grid average filtering. t-SNE, as introduced in \citep{van2008visualizing}, is used to embed the downsampled 3D point cloud into 2D space. Superpoint extraction is then performed exclusively in 2D domain. t-SNE inherently provides clusters that are formed based on the implicit manifolds to which the 3D points belong. The assumption that the data points lie on smooth low-dimensional manifolds accords with the fact that, for many plant species, plant points lie on one or two-dimensional manifolds. t-SNE makes it possible to unwrap the structure of the 3D plant model onto a single 2D view.

The first step of the superpoint extraction scheme corresponds to 2D Euclidean clustering to separate the clusters formed by t-SNE. Line-like structures are detected through examining the local geometry of the points. An iterative procedure that involves spectral clustering and solidity computation is employed to segment non-convex clusters into convex regions. Once all 2D points are assigned to their corresponding superpoints, the assignments are carried to their 3D counterparts, and propagated to the high-resolution point cloud through nearest neighbor interpolation.

In this work, for the PLANesT-3D dataset, the grid size for voxel grid average filtering is selected as 0.12cm prior to t-SNE embedding. Grid size controls the computational load of t-SNE execution and the level of detail that is desired to be transferred to the 2D domain as the preserved local structure. For the plant models in the dataset, we aimed for a resolution close to 1mm while reducing the number of points to be processed by t-SNE. The $Perplexity$ value determines the effective neighborhood in terms of number of data points for t-SNE to calculate similarity. We targeted to have 3 to 4 cm of range for to measure the similarity between two points during the operation of t-SNE. Therefore, given that the grid size is set to 0.12cm, we chose the $Perplexity$ value as 30. The distance threshold used in Euclidean clustering in 2D space is chosen as 1 through the observation that minimum separation between clusters in 2D space is around one unit. The other parameters were kept in their default settings as given in \citep{dutagaci2023using}.

\underline{\emph{Leaf-stem classification network:}} Each superpoint generated by the previous step is considered to be a separate point cloud. A point-based deep neural network, introduced by \cite{turgut2023local}, is utilized to classify the superpoints into leaf and stem classes (Fig. \ref{fig:flowSeg}). The network is formed through integrating two modules to the PointNet++ architecture for object classification. The modules are based on attention mechanisms to analyze point interactions within and between point neighborhoods, and are named as Center Shift Module (CSM) and Radius Update Module (RUM) \citep{turgut2023local}. Their task is to update adaptively the centers and radii of the spherical regions on which PointNet++ encodes local features. This operation is referred to as adaptive local region inference \citep{turgut2023local}. The details of the network are provided in Appendix \ref{sec:app}.

\subsubsection{PointNet++}
\label{sec:pointnet2}

We adopted the PointNet++ architecture \citep{qi2017pointnet2} for semantic segmentation of large point clouds. The data preparation for PointNet++ involves two operations: Subsampling and partitioning the big point set into subsets enclosed by blocks.

The subsampling is performed using voxel-grid filtering, where the edge length of the voxel cube is set as $0.1cm$. Each voxel is represented with a single point which is the average of the original points occupying the voxel. The sampled point is assigned the label of the closest point in the original cloud. This subsampling operation produces a point cloud with reduced size and homogeneous point density. 

3D point-based deep neural networks require input clouds with a fixed number of points. One way is to subsample the full plant point cloud to have the required number of points. However, such strategy will result in significant loss of geometric information, especially for small structures. Instead, we selected to partition the plant point cloud into subsets of points, each of which is then processed by the network individually. We followed a similar procedure of partitioning the point clouds into blocks as in \cite{Turgut2022segmentation,Turgut2022rosesegnet}.

The horizontal region, i.e. the extent of the $XY$ plane, encompassing the point cloud is partitioned into squares of edge length of $10cm$. All points with $x$ and $y$ coordinates falling into a square region is considered belonging to the point subset within a block. The points in the block are subsampled to obtain a cloud of fixed number of points, as $N = 8192$. The partitioning operation is conducted twice for each plant point cloud with offset values $0$ and $5cm$. 
 
We used the default PointNet++ architecture for semantic segmentation as given in \citep{qi2017pointnet2}. PointNet++ segmentation network is composed of five grouping/abstraction layers. The radius parameter, which defines the size of local regions for grouping and abstraction, is set for each of the first four layers as $0.5cm$, $1cm$, $2cm$, and $4cm$, respectively. During training, batch size was selected as 16. The learning rate was updated as 0.005. Other hyperparameters of the PointNet++ were kept at default values.

\subsubsection{RoseSegNet}
\label{sec:rosesegnet}

RoseSegNet was developed for the specific application of semantic segmentation of rosebush models. It involves attention-based modules that encode point interactions within and between local regions. The details of the RoseSegNet architecture can be found in \citep{Turgut2022rosesegnet}.

As with the case of PointNet++, we partitioned the plant point cloud into blocks and the point subset in each block was separately processed by RoseSegNet. The block partitioning process is the same as with PointNet++. 

All the hyperparameters of the RoseSegNet are kept at their default values as given in \citep{Turgut2022rosesegnet}, where the parameter search was performed on the ROSE-X dataset \citep{Dutagaci2020rose}.


\section{Results}
\label{sec:results}

In this section, we provide semantic segmentation results on the PLANesT-3D dataset obtained by the three methods described in Section \ref{sec:threemethods}. We trained a separate network for each of the three plant species. The hyperparameters of the networks were kept fixed for all the species. 70\% of the point clouds were selected for training the networks. The rest is reserved for test. The splits for training and test sets for the three species are given in Table \ref{tab:split}. The IDs of the plants in the test sets are also provided.

\begin{table}[!ht]
  \centering
  \caption{The split for training and test sets}
  \label{tab:split}
  \begin{tabular}{l|ccc}
  Species & Pepper & Rose & Ribes \\ \hline
   \# plants for training & 7 & 7 & 10 \\ \hline
    \# plants for test & 3 & 3 & 4 \\ \hline
     Plant IDs reserved for test & 01, 03, 07 & 01, 03, 09 & 03, 10, 11, 14 \\ \hline
     \end{tabular}
\end{table}

Before being processed by the networks, the point clouds were filtered such that points with confidence values less than 6 were discarded. For PointNet++ and RoseSegNet, the point cloud of a plant is partitioned into vertical blocks of size $10cm \times 10cm$ of base area, with offset values of 0 and $5cm$. The points in each block are subsampled to obtain $N=8192$ points per block. Each block is processed by the PointNet++ and RoseSegNet independently as a separate point cloud, as described in Section \ref{sec:pointnet2}. The number of blocks extracted from the training and test plant models are given in Table \ref{tab:blocks}. The training blocks correspond to the blocks extracted from the plant point clouds reserved for training the networks.

For SP-LSCnet, the point clouds are partitioned into superpoints, and the superpoints are treated as separate objects for the object classification network. Table \ref{tab:blocks} gives the number of superpoints extracted from the training and test plant models for each plant species. 

\begin{table}[!ht]
  \centering
  \caption{Number of point subsets used for training and test.}
  \label{tab:blocks}
  \begin{tabular}{l|ccc}
  Species & Pepper & Rose & Ribes \\ \hline
   \# training blocks for PointNet++ \& RoseSegNet & 294	 & 317 & 665\\ \hline
   \# test blocks for PointNet++ \& RoseSegNet & 123 & 106 & 205 \\ \hline
   \# training superpoints for SP-LSCnet & 1188 & 933  & 2761 \\ \hline
   \# test superpoints for SP-LSCnet & 646  & 438 & 998 \\ \hline
  \end{tabular}
\end{table}

\begin{figure*}[!ht]
\centering
\includegraphics[width=0.8\textwidth]{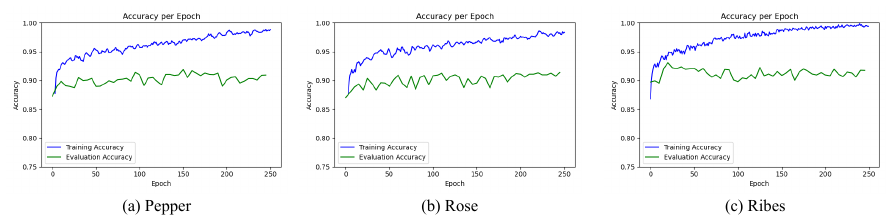}
\caption{Training curves for the classification network LSCnet for the three species. Accuracy corresponds to the classification accuracy of superpoints in training and evaluation (test) sets.}
\label{fig:trainingCurves}
\end{figure*}

\begin{figure*}[!ht]
\centering
\includegraphics[width=0.7\textwidth]{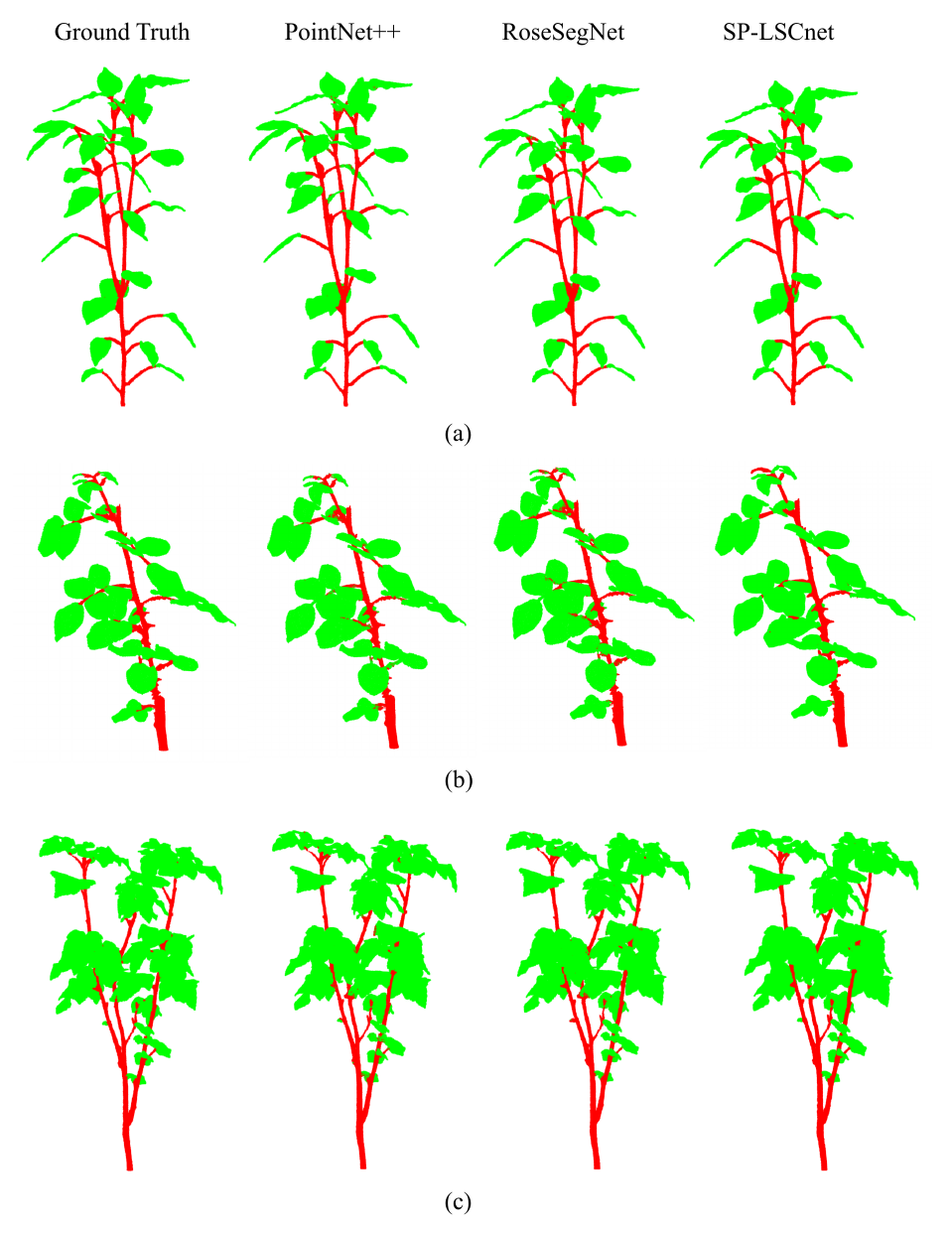}
\caption{Semantic segmentation results by PointNet++, RoseSegNet, SP-LSCnet, on sample pepper (a), rose (b), and, ribes (c) plants. First column corresponds to the ground truth. }
\label{fig:visResults}
\end{figure*}

The classification network LSCnet takes superpoints as individual point clouds as input and trains the network to classify the superpoints into "leaf" or "stem" objects. A separate model was trained for each of the three species. Fig. \ref{fig:trainingCurves} gives the training curves for the classification network LSCnet for the three species. Accuracy corresponds to the classification accuracy of the superpoints in training and evaluation (test) sets. Maximum number of epochs was set to 250. The accuracy on the training set continued to increase steadily. The accuracy on the test set kept increasing, although slightly and with oscillations, until 150th epoch for the pepper plants and until 250th epoch for the rose plants. For the ribes plants, the performance on the test set did not improve further after 20 epochs. However, the oscillations did not fall below the accuracy rate of 0.9 in the subsequent epochs. These results indicate that the LSCnet did not suffer from overfitting during the training process.

\begin{table}[!th]
  \centering
  \caption{Semantic segmentation results of the three methods on the PLANesT-3D dataset.}
  \label{tab:results}
  \small
  \begin{tabular}{r|ccc}
  \hline
   &  & \textbf{PEPPER} &  \\ 
   & PointNet++ & RoseSegNet &  SP-LSCnet\\ \hline
Precision - Stem &	96.5	& 97.0	& 97.6 \\ \hline
Recall - Stem &	94.6	& 96.1	& 94.5 \\ \hline
IoU - Stem &	91.5	& 93.2 &	92.4 \\ \hline
Precision - Leaf &	98.3	& 98.7	& 98.3 \\ \hline
Recall - Leaf	& 98.9	& 99.0 &	99.3 \\ \hline
IoU - Leaf &	97.2	& 97.8 &	97.6 \\ \hline
Acc	& 97.9	& 98.3 &	98.1 \\ \hline
MIoU	& 94.3 &	95.5 &	95.0 \\ \hline

   &  & \textbf{ROSE} &  \\ 
   & PointNet++ & RoseSegNet &  SP-LSCnet\\ \hline
   Precision - Stem &	92.4	& 94.8	& 96.4 \\ \hline
Recall - Stem &	89.7	& 91.4	& 85.6 \\ \hline
IoU - Stem	& 83.5	& 87.1	& 83.0 \\ \hline
Precision - Leaf &	97.6	& 98.0	& 97.2 \\ \hline
Recall - Leaf	& 98.3	& 98.8	& 99.4 \\ \hline
IoU - Leaf	& 96.0	& 96.9	& 96.6 \\ \hline
Acc	& 96.7	& 97.4	& 97.1 \\ \hline
MIoU	& 89.8	& 92.0	& 89.8 \\ \hline
   &  & \textbf{RIBES} &  \\ 
   & PointNet++ & RoseSegNet &  SP-LSCnet\\ \hline
   Precision - Stem	& 95.8	& 95.9	& 96.7 \\ \hline
Recall - Stem	& 94.0	& 95.5	& 94.0 \\ \hline
IoU - Stem	& 90.3	& 91.7	& 91.1 \\ \hline
Precision - Leaf &	98.9	& 99.2	& 98.5 \\ \hline
Recall - Leaf	& 99.2	& 99.2	& 99.2 \\ \hline
IoU - Leaf	& 98.1	& 98.4	& 97.8 \\ \hline
Acc	& 98.4	& 98.6	& 98.3 \\ \hline
MIoU	& 94.2	& 95.1	& 94.5 \\ \hline
   
  \end{tabular}
\end{table}

\subsection{Quantitative results}

We report the semantic segmentation results of the three methods in Table \ref{tab:results}. Precision, Recall, and Intersection over Union (IoU) measures specific to "stem" and "leaf" classes are given in the table, as well as overall accuracy (Acc) and mean IoU (MIoU). In terms of overall accuracy and mean IoU, RoseSegNet yielded the best performance for all the three species. In terms of MIoU, RoseSegNet surpassed the performance of PointNet++ by close to 1\%. SP-LSCnet is in between, performing slightly better than PointNet++ for the Pepper and Ribes sets, while yielding similar peformance to that of PointNet++ for the Rose set.

\begin{figure*}[!t]
\centering
\includegraphics[width=0.3\textwidth]{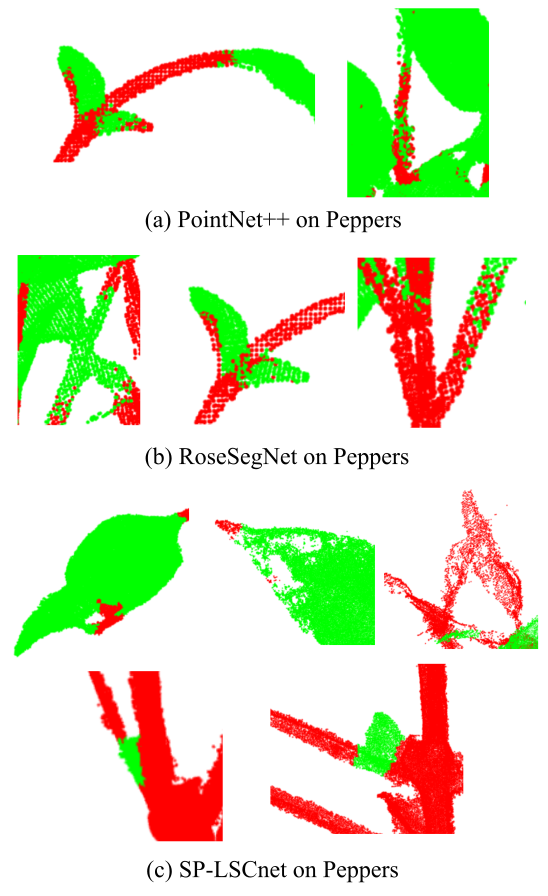}
\caption{Examples of errors of the three methods on Pepper point clouds.}
\label{fig:errorsPeppers}
\end{figure*}

\begin{figure*}[!t]
\centering
\includegraphics[width=0.3\textwidth]{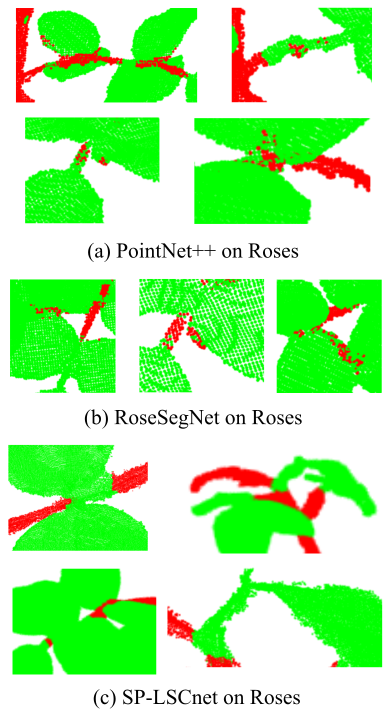}
\caption{Examples of errors of the three methods on Rose point clouds.}
\label{fig:errorsRoses}
\end{figure*}

\begin{figure*}[!t]
\centering
\includegraphics[width=0.3\textwidth]{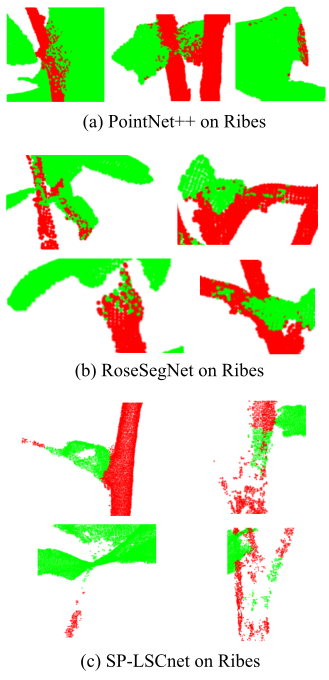}
\caption{Examples of errors of the three methods on Ribes point clouds.}
\label{fig:errorsRibes}
\end{figure*}

The segmentation performance figures on rose plant models are lower compared to those obtained on pepper and ribes plants. The recall of the "stem" class is more challenging for the rose plant models, since the petioles in between the leaflets are hard to discern. For other plants, the MIoU figures are still below 96 \%, indicating that there is room for improvement. The following subsection, where a visual error analysis is given, provides more insight into the causes of errors.

\subsection{Visual results}

Fig. \ref{fig:visResults} gives visual semantic segmentation results by PointNet++, RoseSegNet, SP-LSCnet on sample point clouds. In Figs. \ref{fig:errorsPeppers}, \ref{fig:errorsRoses}, and \ref{fig:errorsRibes} examples to erroneous regions produced by the three methods are demonstrated for pepper, rose, and ribes plants, respectively.  For the rose plants, all three methods produce erroneous results at thin petioles where the opposing leaflets are very close (Fig. \ref{fig:errorsRoses}). The misclassified points of PointNet++ and RoseSegNet are observed as speckles. With SP-LSCnet, the erroneous regions are smooth as they were misclassified together with the superpoint they belong to.

The methods also produce errors at leaves and other stem parts of the plants; however the nature of the errors of SP-LSCnet is different from those of PointNet++ and RoseSegNet. With the laters, the errors appear as isolated points scattered in the regions (Figs. \ref{fig:errorsPeppers}, \ref{fig:errorsRoses}, and \ref{fig:errorsRibes}). Although context information heavily influences the features of single points in PointNet++ and RoseSegNet, the classification is performed individually for each point. Misclassifications occur particularly at cluttered regions where leaves and stems are close to each other.

The treatment of each superpoint as a single entity by SP-LSCnet produces smooth results; however, it suffers from two disadvantages: First, the superpoints might not be always homogeneous; some leakage from other classes is inevitable due to the unsupervised nature of our approach for superpoint extraction. Second, the correct classification of the superpoints becomes particularly critical. At the junctions, portions of stems may be separated from the adjacent branches and misclassified as leaves, as observed in Fig. \ref{fig:errorsPeppers}c and Fig. \ref{fig:errorsRibes}c. In these cases, the stem points were classified as leaf points since the single superpoints to which they belong have the geometrical characteristics of leaves as modeled by the network. Likewise, small leaves, where the point resolution is not sufficient for proper sampling of the leaf surface, are classified as stem (Fig. \ref{fig:errorsPeppers}c).

PointNet++ and RoseSegNet have a tendency to misclassify borders of leaves as stems as can be obserfed in Figs. \ref{fig:errorsPeppers}, \ref{fig:errorsRoses} and \ref{fig:errorsRibes}. Such errors are rare with SP-LSCnet since borders of a leaf are not partitioned into separate superpoints. However, when there is a hole in the leaf or the point density drops at a certain region, as are the two cases in the leaves shown in Fig. \ref{fig:errorsPeppers}(c), parts of the leaves can be separated into different superpoints and misclassified as stems.

An important cause of error of the SP-LSCnet is low point density and missing structures due to self occlusion. Regions of low point density can cause isolated regions separated into superpoints which are not large enough to possess distinguishing shapes. Three examples of such errors are illustrated in Fig. \ref{fig:errorsRibes}(c).

All three methods can benefit from post-processing such as label smoothing for PointNet++ and RoseSegNet, and graph-based re-evaluation of superpoint predictions for SP-LSCnet. However, the limitations of these methods in dealing with scarce annotated data, noisy point clouds, complex architecture of plants, and variable point density call for further developments in plant segmentation methods.

\section{Discussion}
\label{sec:discussion}
The main outcome of this study is the new PLANesT-3D dataset, which is composed of 3D point clouds of plants reconstructed through structure from motion and multi-view stereo. Capturing 3D geometry through SfM has many advantages such as low cost setup requirements, high resolution, and provision of color information \citep{harandi2023make}. It can be used both in controlled environments and in the field without extensive training for acquisition and elaborate protocols. A public dataset obtained through this modality as opposed to laser scanning is expected to be instrumental in assessing the robustness of 3D plant processing and phenotyping tools.

PLANesT-3D is small in terms of the number of point clouds (34 point clouds) as compared to Pheno4D (126 point clouds) and Soybean-MVS (102 point clouds). It also lacks temporal data, i.e., all plants in PLANesT-3D were captured once and were not tracked during their development as opposed to the plants in Pheno4D (14 plants were tracked) and Soybean-MVS (5 plants were tracked). However, each of the point clouds of PLANesT-3D corresponds to a distinct plant with diverse overall shapes, architectures, and organ shapes and texture.

Obtaining 3D geometry through SfM has its limitations. One is the necessity of acquiring sharply-focused images from multiple locations. Each point on the object surface should be observed from at least two camera viewpoints for complete coverage. The manual acquisition process followed in this study may fall below the desired speed for high-throughput phenotyping. Furthermore, SfM is computationally intense compared to 3D geometry capture from Laser scanners or depth cameras, especially for generating dense, high-quality point clouds from large sets of images.

We used quite large numbers of images of the plants for reconstruction (\ref{tab:numimages}). This amount of image capture might not be feasible in cases where manual data acquisition is performed. However, advances in robotics enable fast acquisition 
with autonomous crop surveying robots equipped with robotic arms and multiple cameras \citep{atefi2021, iqbal2020development, smitt2021pathobot}, as well as techniques for automatic positioning of the robot arms for optimal data acquisition \citep{wu2019plant}. Nevertheless, limiting the number of images for reconstruction without a performance degradation in particular tasks would substantially reduce the computational demand.

We provided the 2D color images that were used to reconstruct the plant models as part of the PLANesT-3D dataset. As part of a future scenario, these images can be used to test new 3D reconstruction techniques that use 2D photographs as input, in terms of quality and computational efficiency. 

As can be observed in Table \ref{tab:properties}, the size of each model in terms of number of points is high, ranging from 600K to 5.5M points. One question is whether such high resolution is necessary for certain tasks related to plant analysis and management. For all the semantic segmentation techniques used in this study, the number of points were reduced through downsampling as a preprocessing step. As future work, the downsampling rate can be varied systematically to measure the effect of resolution on the performance of particular 3D shape processing tools \citep{Chaudhury2021transferring}. Another possibility is to search for optimal subsets of the images we provided for 3D reconstruction to reduce point density without loss of significant geometric information due to self-occlusion.

A weakness of this study is the lack of fully exploring the potential contribution of the color despite the availability of color information in the PLANesT-3D point clouds. The use of color information together with 3D geometry can boost the performance of a segmentation procedure \citep{Boogaard2021boosting}. The potential boost depends on the plant species. For some plants, leaves and branches are indistinguishable in terms of color, for others color is one of the dominant attributes. An analysis on the contribution of color information deserves an extensive study. In this future analysis, the use of color correction schemes, conversion to various color spaces, and effects of illumination and plant species should be thoroughly explored.

The 3D color point clouds in PLANesT-3D, together with the corresponding color images, will form a valuable dataset, for which the acquisition was performed under uncontrolled illumination conditions. Also, color bleeding from the background in areas where there are abrupt changes in depth, i.e. boundaries of the organs, is present in the point clouds of the PLANesT-3D dataset. This issue needs to be resolved by methods aiming to integrate color information into the segmentation process.

We limited the study to the segmentation of isolated plants into their semantic parts. The evaluated techniques were set to operate under the condition that the plant point cloud was separated from the background through a prior step, and the correct scale of the plant was recovered. Segmentation of plants in cluttered environments, where object occlusions are also present, is beyond the scope of this study. However, we provide, as part of PLANesT-3D, the raw point clouds produced by Metashape Professional; i.e. the original point cloud prior to the application of our semi-automatic plant cropping and scaling procedure. These point clouds can be used to test techniques specifically designed to automatically separate plant points from the cluttered background, to recover correct scale, and to provide pose normalization.

Semantic parts annotated in the PLANesT-3D dataset are limited to "leaf" and "stem" classes; which we believe is another limitaion of this study. A more fine-grained annotation is possible, for example, in terms of "leaf blades", "petioles", "main stem", "nodes", and "internodes" \citep{Boogaard2022}. Leaflets, together with petioles, can be grouped into their corresponding individual leaves. The architecture of each plant can be modeled with a graph representation. We plan to undertake this level of annotation as a future work.

Another limitation of this study is the lack of application and comparison of panoptic segmentation algorithms, despite the availability of instance labels of the leaves. SP-LSCnet has the potential of being modified as an instance and panoptic segmentation method since it already performs a clustering of 3D points in its first stage. Once semantic labels are predicted in the second stage, these clusters can go through a proximity and homogeneity analysis for being merged into individual leaf instances. In a future study, the performance of this version of LSCnet can be compared with state-of-the-art panoptic segmentation networks extensively on PLANesT-3D datset and other datasets providing instance labels such as Pheno4D and Soybean-MVS.

We can list three more outcomes of this study as follows: 1) The semantic segmentation results on the new dataset obtained through the widely-used PointNet++ will serve as baselines for future studies. 2) RoseSegNet, without the necessity of tuning its hyperparameters, was able to surpass the segmentation performance of PointNet++. 3) SP-LSCnet, which is an alternative to end-to-end networks, provided competitive segmentation results. The MIoU results for each of the plant species were close with the three methods: around 95\% for Pepper and Ribes, and around 90\% for Rose. 

The accuracy of semantic segmentation is particularly important, since semantic segmentation is either a prior step (\citep{GUO2023}) or a parallel branch or module (\citep{Vu2024,Li2022}) to instance or panoptic segmentation procedures. There is a significant imbalance between the number of points corresponding to the leaf and stem points due to the nature of the plants. The IoU for leaf points are high while, the IoU for stem is below 93\% with all the methods. Correctly identifying the branches is particularly important for both panoptic segmentation and modeling the architecture of the plant. There is room for improvement in terms of both the precision and recall of the stem points, especially for the rose plants.

Furthermore, with networks trained with other datasets of the same species, or with models from different species, one expects that the semantic segmentation results will drop below the baseline. The semantic segmentation results provided in this study, where a separate network was trained for each of the three plant species, can represent the achievable baselines for future studies on the transferability of the neural network models. The extent of generalization ability of network models among plant species is of special interest; since manual annotation of plants for training is highly labor-intensive. 

{In this regard, PLANesT-3D is important to the community in terms of species diversity, with a collection of three different species. The low-cost acquisition modality also contributes greatly to the data variation that will allow to evaluate algorithms for  reproducibility, generalization ability, and suitability for transfer learning, and small sample learning.

This study also involves the introduction of the new technique, SP-LSCnet, which is composed of two modules. The first module, through t-SNE, extracts clusters of the 3D point clouds while visualizing the clusters in a single 2D layout. This visualization provides interpretability to the pipeline, as opposed to end-to-end deep learning architectures that operate as black-boxes. The second module can be replaced by any object classification network that processes 2D or 3D point clouds; allowing the method to be updated by more efficient methods in the future. The pipeline is also suitable for an interface where the user can visualize misclassified clusters in 2D and correct them if necessary.

Although PointNet++, RoseSegNet and SP-LSCnet seem to be performing similarly in terms of quantitative measures, the errors they produce have different natures (Figs. \ref{fig:errorsPeppers}, \ref{fig:errorsRoses}, and \ref{fig:errorsRibes}). PointNet++ and RoseSegNet, in many cases, misidentify borders of leaves as stems. With SP-LSCnet this type of error is rare, as explained in Section \ref{sec:results}. Both methods tend to misclassify thin petioles between close leaflets of the roses; however, while PointNet++ and RoseSegNet's errors are scattered individual points, SP-LSCnet misclassifications appear as smooth regions in between leaflets. Other sources of error for SP-LSCnet are the misidentification of small clusters produced by t-SNE at branch junctions and regions of low point density. Including context information using multi-scale clustering with varying perplexity values, which are then organized in a graph structure can help reduce such errors of the SP-LSCnet framework.

One weakness of SP-LSCnet is the reliance on the semantic homogeneity of superpoints, which is not always guaranteed, especially at leaf-stem boundaries. Another, as mentioned above, is the loss of context, since each superpoint is processed individually. Our future work involves remedying these weaknesses by processing multiple t-SNE maps in a hierarchical manner, organizing superpoints in graph structures, and using graph networks for classifying each node into its semantic class. 

The effectiveness of SP-LSCnet, should also be tested on other 3D plant datasets mentioned in Table \ref{tab:datasets}. Furthermore, cross-validation tests should be performed, where the network is trained on one dataset (or species) and tested on another. In order to keep this current study focused, an extensive cross-validation study of the current SP-LSCnet and its improved versions is reserved to be part of a future work.

\section{Conclusions}
\label{sec:conclusions}

We introduced the PLANesT-3D dataset, which is composed of annotated color point clouds of 34 plants belonging to three different species. We described the acquisition, reconstruction, and labeling processes involved in dataset construction. PLANesT-3D will be a valuable benchmark for assessing the generalization ability and reproducibility of current and future 3D plant phenotyping tools on diverse data. PLANesT-3D includes new instances from three different species. Point clouds of two of these species are provided for the first time. The instances posses a large within-species variety in terms of plant size, architectural complexity, and the number, density, geometry and texture of organs. Also, the new dataset is constructed with the low-cost SvM/MVS technique, introducing challenges such as varying point density, missing data, reconstruction noise, and color bleeding.

As a use case of the dataset, we reported semantic segmentation results yielded by PointNet++, RoseSegNet, and a novel method abbreviated as SP-LSCnet. RoseSegNet achieved highest segmentation performance, reaching 95.5 \%, 92.0 \%, and 95.1 \% MIoU, for pepper, rose, and ribes plants, respectively. SP-LSCnet also produced comparable results as 95.0 \%, 89.8 \%, and 94.5 \% MIoU on the three plant species. The main advantage of SP-LSCnet is its two-stage structure, where the first stage is an unsupervised scheme that provides 2D visualization and the second stage can be replaced by any classification network. The 2D visualization greatly increases the interpretability of the method in comparison to state-of-the-art black-box techniques.

We illustrated visual errors of the three methods in their semantic segmentation outputs. The PLANesT-3D dataset will be instrumental for developing new semantic segmentation methods that effectively address the challenges illustrated by these errors. As a dataset containing plant point clouds of diverse structures with regions of varying quality, it will help assessing tools for 3D plant analysis tasks, such as panoptic segmentation, leaf counting, leaf shape characterization, skeletonization of the branch structure, decoding of the architecture, which are essential for automating plant research and agricultural operations.

\section*{Acknowledgements:}
The authors acknowledge the support of The Scientific and Technological Research Council of Turkey (TUBITAK), Project No: 121E088.

\section*{Declaration of Competing Interest}
The authors declare that they have no known competing financial interests or personal relationships that could have appeared to influence the work reported in this paper.

\appendix
\section{LSCnet}
\label{sec:app}

In this section, we provide details on the Leaf-Stem-Classification network (LSCnet), which classifies individual superpoints (Fig. \ref{fig:flowSeg}). The network is formed through integrating the Center Shift Module (CSM) and the Radius Update Module (RUM) \citep{turgut2023local} to the PointNet++ architecture for object classification (Fig. \ref{fig:pointnet2}). The modules update adaptively the centers and radii of the spherical regions on which PointNet++ encodes local features.

\begin{figure}[!ht]
    \centering
    \begin{subfigure}[b]{1.0\textwidth}
         \centering
         \includegraphics[width=\textwidth]{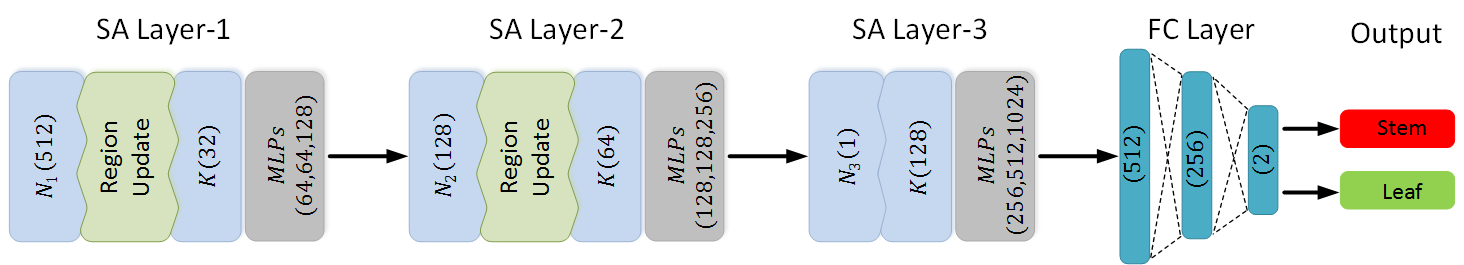}
         \caption{}
         \label{fig:poitnet2_network}
    \end{subfigure}
    \begin{subfigure}[b]{0.8\textwidth}
         \centering
         \includegraphics[width=\textwidth]{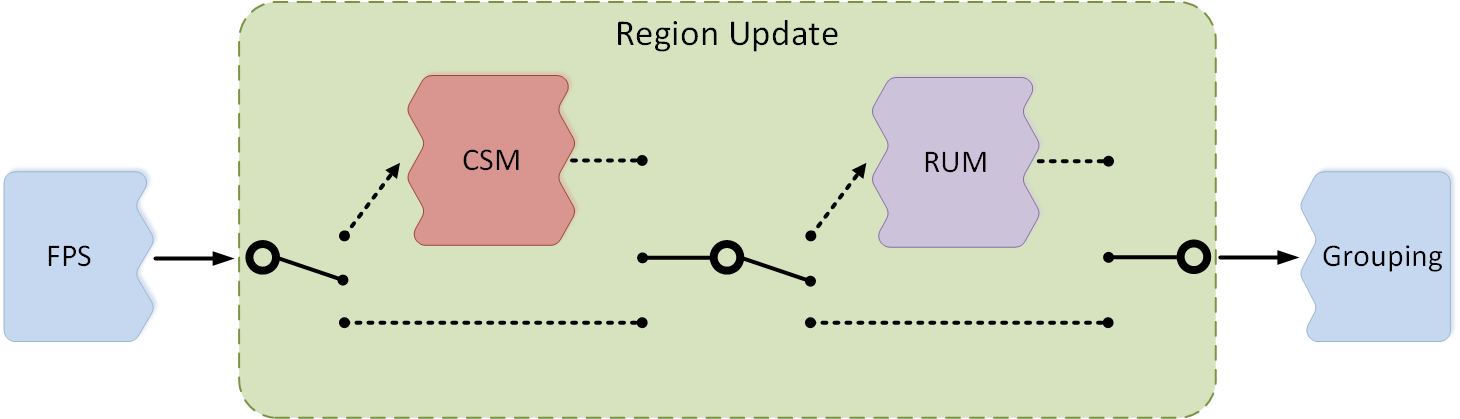}
         \caption{}
         \label{fig:poitnet2_modules}
     \end{subfigure}
     \caption{The integration of CSM and RUM to PointNet++ framework \citep{turgut2023local}: The classification network. $r^L$ for $L = 1$ (first layer) is set to be 0.2. $r^L$ for $L = 2$ (second layer) is 0.4 (a).    Adaptive local region inference with CSM and RUM (b).
     \label{fig:pointnet2}}
\end{figure}

The classification network architecture of PointNet++ \citep{qi2017pointnet2}, shown in Fig. \ref{fig:poitnet2_network}, includes three set abstraction (SA) layers. The input point cloud with $N = 1024$ points is reduced to $N_1 = 512$, and $N_2 = 128$ representative points during SA layer-1 and SA layer-2, respectively. At each SA layer $L$, the representative points $c_j^L$ are selected using farthest point sampling (FPS) algorithm. $N_L$ spherical local regions with radius $r^L$ is defined around the center points $c_j^L$. Within each region, $K$ points are randomly selected and the features of $K$ points are mapped to higher dimensions with MLPs. The abstracted features of each local region are calculated by taking the maximum among the feature channels of $K$ grouped points.  

The global descriptors representing the input point cloud are extracted at the last layer, where a single region enclosing the entire object is encoded. The global descriptors are passed through fully connected layers to compute the category scores. In this study, the classification network calculates two category scores for each input point cloud: One for the "leaf" category, and one for the "stem" category.

The adaptive local inference modules, CSM and RUM can be integrated to either or both of the first two layers as shown in Fig. \ref{fig:poitnet2_modules}. In other words, CSM and RUM modules can be turned "ON" or "OFF" at particular layers of the network. If CSM module is set as "ON", the representative points are updated. If it is "OFF", representative points remain as the ones computed by the FPS algorithm. Likewise, if the RUM module is "ON", the radius around each representative point is altered, otherwise it is set to the constant value specific to each layer. In this way, the user can control at which layers the receptive fields are altered adaptively using attention mechanisms.

If CSM is set as "ON" at a particular layer, for each representative point $c_j^L$, CSM calculates the shift amount $\Delta c_j^L$, through attention-based subnetworks. The representative point is updated as:

\begin{equation} \label{eq:hatcj}
   \hat{c}_{j}^{L} = c_{j}^{L} + \Delta c_{j}^{L}\;.
\end{equation}

If RUM is "ON", it computes the amount of radius update $\Delta r_{j}^{L}$ for each region $j$. The new radius then becomes:

\begin{equation} \label{eq:radius}
   \hat{r}_{j}^{L} = r^{L} + \Delta r_{j}^{L}\;.
\end{equation}
The grouping and feature encoding by MLPs of $K$ points are then performed using these new region definitions. These modules allow the receptive fields to be adaptively shifted and resized through examining local and global point interactions through attention. The network thus adjusts the receptive fields and encodes the 3D points in them in accordance with the main task of the network.

Various alternatives of CSM and RUM were suggested by \cite{turgut2023local}. As a result of experimental trials, the CSM-II (sub) and RUM-II (cum) versions were selected for the PLANesT-3D dataset. Please, refer to \citep{turgut2023local} for detailed descriptions of the modules CSM-II (sub) and RUM-II (cum).

\bibliographystyle{elsarticle-num-names}
\bibliography{ref.bib}
\end{document}